\documentclass[12pt,draftclsnofoot,onecolumn]{IEEEtran}
\usepackage{booktabs}
\usepackage{graphicx}
\usepackage[cmex10]{amsmath}
\usepackage{cases}
\usepackage[tight,footnotesize]{subfigure}
\usepackage{amsthm} 
\usepackage{amsmath}
\usepackage{cite}
\usepackage{amssymb}
\usepackage{algorithm}
\usepackage{algorithmic}
\usepackage{xcolor}
\usepackage{stfloats}
\usepackage{multirow}
\usepackage{CJK}
\usepackage{subeqnarray}
\usepackage{longtable}
\usepackage{multicol}
\usepackage{extpfeil}
\usepackage{verbatim}

\ifCLASSINFOpdf
\else
\fi

\begin{document}
%
\title{Phase Shift Design in RIS Empowered Wireless Networks: From Optimization to AI-Based Methods }


\author{Zongze Li, Shuai Wang, Qingfeng Lin, Yang Li, Miaowen Wen, Yik-Chung Wu, \\and H. Vincent Poor
	
\thanks{Zongze Li is with Peng Cheng Laboratory, Shenzhen 518038, China (e-mail: lizz@pcl.ac.cn). Qingfeng Lin and Yik-Chung Wu are with the Department of Electrical and Electronic Engineering, The University of Hong Kong, Hong Kong (e-mail: qflin@eee.hku.hk; ycwu@eee.hku.hk).
Shuai Wang is with the Shenzhen Institute of Advanced Technology, Chinese Academy of Sciences, Shenzhen 518055, China (e-mail: s.wang@siat.ac.cn).
Yang Li is with Shenzhen Research Institute of Big Data, Shenzhen 518172, China (e-mail: liyang@sribd.cn).
Miaowen Wen is with the School of Electronics and Information Engineering, South China University of Technology, Guangzhou 510640, China (e-mail: eemwwen@scut.edu.cn). 
H. Vincent Poor is with the Department of Electrical and Computer Engineering, Princeton University, Princeton, NJ 08544 USA (e-mail: poor@princeton.edu).}
}


\maketitle

\begin{abstract}
Reconfigurable intelligent surfaces (RISs) have a revolutionary capability to customize the radio propagation environment for wireless networks. To fully exploit the advantages of RISs in wireless systems, the phases of the reflecting elements must be jointly designed with conventional communication resources, such as beamformers, transmit power, and computation time. However, due to the unique constraints on the phase shift, and massive numbers of reflecting units and users in large-scale networks, the resulting optimization problems are challenging to solve. This paper provides a review of current optimization methods and artificial intelligence-based methods for handling the constraints imposed by RIS and compares them in terms of solution quality and computational complexity. Future challenges in phase shift optimization involving RISs are also described and potential solutions are discussed.
\end{abstract}


%
\IEEEpeerreviewmaketitle

\section{Introduction}

It is well-known that line-of-sight (LoS) propagation is a desirable but rarely occurring scenario for wireless communications. Traditional techniques for addressing this issue is to deploy more active nodes such as base stations (BSs), access points, or relays to improve coverage and compensate the high propagation loss in a non-LoS environment. However, this approach would induce high energy consumption and deployment/backhaul/maintenance cost. Worse still, it would also cause more severe and complicated network interference issue. 

Recently, reconfigurable intelligent surfaces (RISs), which are passive devices equipped with large numbers of low cost reflective elements, have emerged as a promising technology to overcome the above challenges. Compared with the conventional active nodes approach which actively transmits the signals, an RIS shapes the incoming signal by adjusting the phase shifts of the reflecting elements. Thus, deploying RISs is more energy-efficient, environmentally friendly, and most importantly free of noise amplification and self-interference~\cite{J_Renzo20RISvsRelay}. Intuitively, deploying an RIS could provide virtual LoS links between a BS and mobile users even when the direct LoS path is blocked by high-rise buildings. Therefore, RISs have significant potential in enhancing both spectral and energy efficiencies in urban environments~\cite{J_Huang19RIS_EE}. Furthermore, due to the passive nature of RISs, they can be flexibly deployed in building facades, indoor walls, aerial platforms, roadside billboards, vehicle windows, etc.

\begin{figure*}	
	\centering
	\includegraphics[scale=0.26]{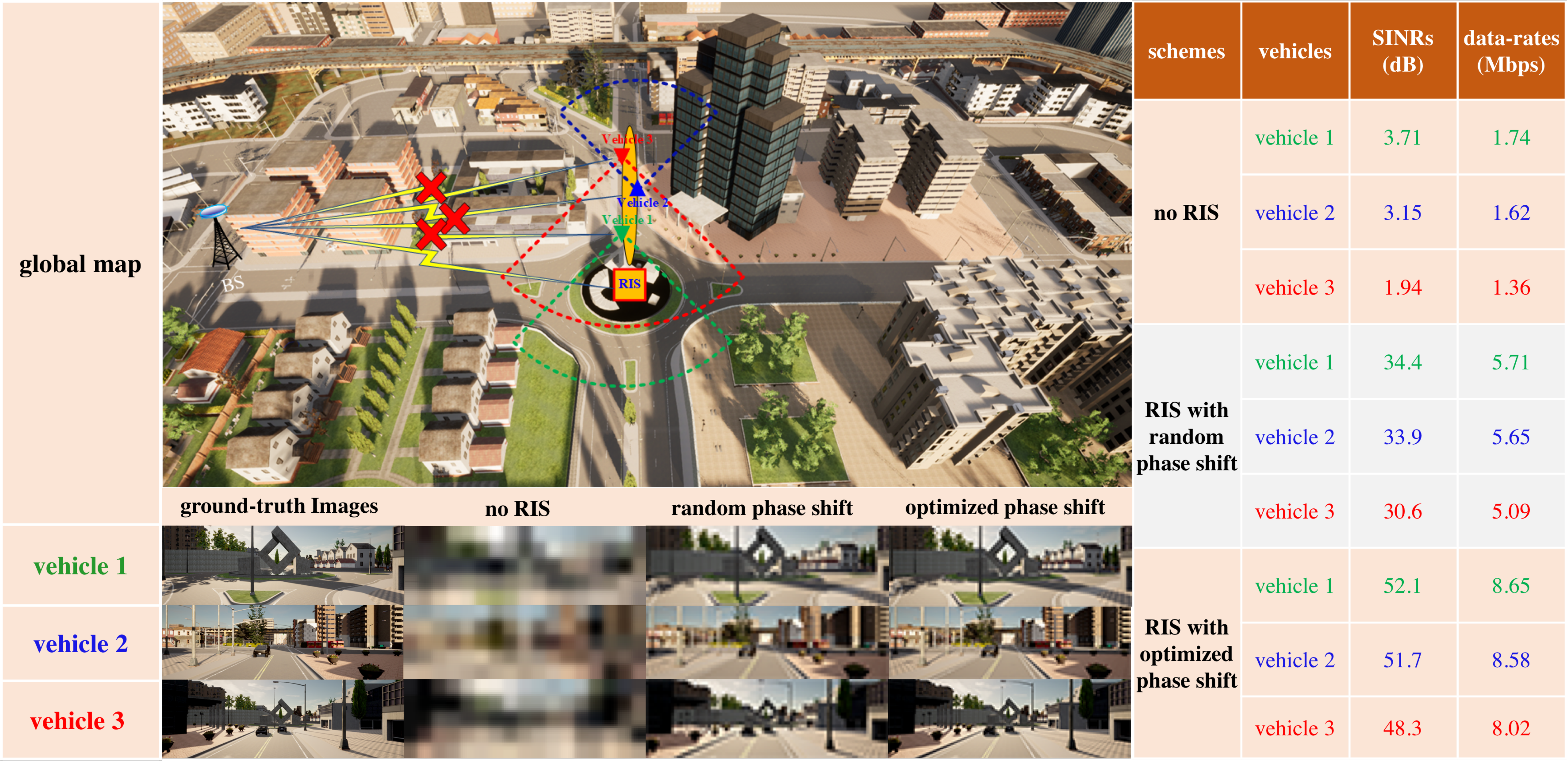}
	\caption{RIS aided V2X for autonomous driving with camera video stream transmission.}\label{alg:V2Xmap_demo}
\end{figure*}

While RISs could be game-changing, their deployment also brings challenges. One of them is resource allocation, which requires the nonconvex constrained phase shifts to be optimized together with other communication resources.  To illustrate the importance of optimizing the phase shifts, we consider a use case on the vehicle-to-everything (V2X) system in Fig.~\ref{alg:V2Xmap_demo}, which consists of a BS located on the left side of the map, an RIS located at the intersection, and three intelligent vehicles marked in different colors. Each car is equipped with a front camera and LiDAR that capture data from the environment. These sensed data need to be transmitted to the BS for cooperative perception, remote driving, or vehicle platooning.  
Due to significant shadowing effects, the received signal power reduces quickly with distance away from the intersection, and high data rate transmission could not be achieved. One can either take a longer duration for transmission, which is not desirable as outdated data is not useful in an intelligent traffic system, or use lossy compression to reduce the amount of data to be sent, which would unfortunately compromise the integrity of information if the compression loss is too much. We illustrate the consequences of the latter option and show how an RIS might help to mitigate them.

In particular, we use the simulation platform of Car Learning to Act (CARLA) and Pytorch in Ubuntu 18.04 with a GeForce GTX 1080GPU for graphic rendering and generation of vivid sensing data~\cite{J_Doso17CARLA}. 
The ground-truth images of a particular frame from the front cameras are shown at the lower-left of Fig.~\ref{alg:V2Xmap_demo}.  
We simulate three transmission schemes: a) direct transmission without an RIS; b) RIS aided transmission with random phase shifts; and c) RIS aided transmission with optimized phase shifts. The signal-to-interference-plus-noise ratios (SINRs) and available data-rates of the three vehicles are shown on the right side of Fig.~\ref{alg:V2Xmap_demo}. Due to the aggressive compression for fitting the data into a poor channel, the images received without the help of an RIS are very blurry.  With an RIS, there is an observable improvement even just with random phase shifts. If the phase shifts of the RIS are optimized, the received images match the ground-truth very well. This demonstrates the necessity of deploying an RIS and the optimization of phase shifts in this V2X communication scenario.

To optimize the non-convex constrained phase shifts at an RIS, a number of optimization methods have been proposed in the literature, including semidefinite relaxation (SDR), the penalty method, the majorization-minimization (MM) algorithm~\cite{J_Sun17MM}, the manifold method~\cite{B_AbsilP09}, gradient descent (GD)~\cite{J_ma20lowcomplexity}, and convex relaxation (CR)~\cite{J_Chen19RISSecure}. Artificial Intelligence (AI) methods, such as unsupervised learning~\cite{J_Gao20unsupervisedLRIS}, supervised learning~\cite{J_TahaDL_RIS}, and reinforcement learning~\cite{J_Huang20RIS_DRL}, also recently emerged as viable solutions. However, the properties of these diverse algorithms are scattered in the literature, and there is a lack of comparisons among them in the context of RISs. To fill this gap, this paper summarizes these techniques, reveals their relationships, and compares their properties. 
	
\section{RIS Resource Allocation Examples and General Formulation}\label{Sec:II}
In wireless resource allocation involving an RIS, there are two types of resources. One is the conventional communication resources, such as beamforming vector, artificial noise, transmit power, and computation time. The other is the RIS coefficients. Each type of resources would have its own constraint, and there are possibly additional constraints coupling the two types of resources. Below are three application examples and their problem formulations. In each of the examples, it is assumed that there are $M$ reflecting elements, and the RIS coefficients are expressed in a vector $\mathbf{e}:=[e_1,\ldots,  e_M]^H \in \mathcal{F}$, with $\mathcal{F}$ being the feasible set of the RIS coefficients, and the specific form of $\mathcal{F}$ will be discussed after the three examples.
\begin{figure*}	
	\centering
	\subfigure[]{ 
		\label{fig:Se_EE_Power1}
		\includegraphics[width=2.9in]{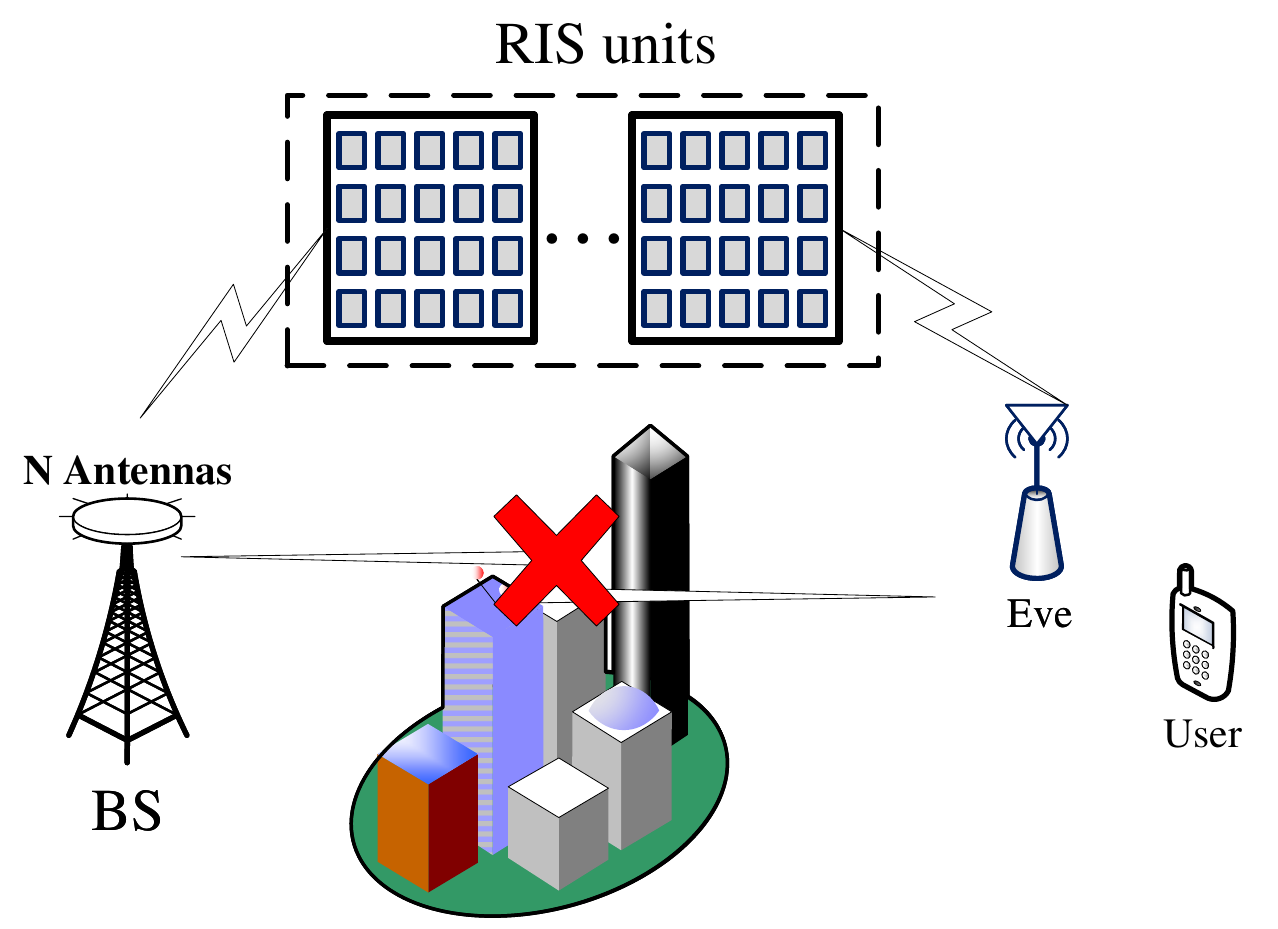}} \hspace{0.0in} 
	\subfigure[]{ 
		\label{fig:UPower_user2}
		\includegraphics[width=3.1in]{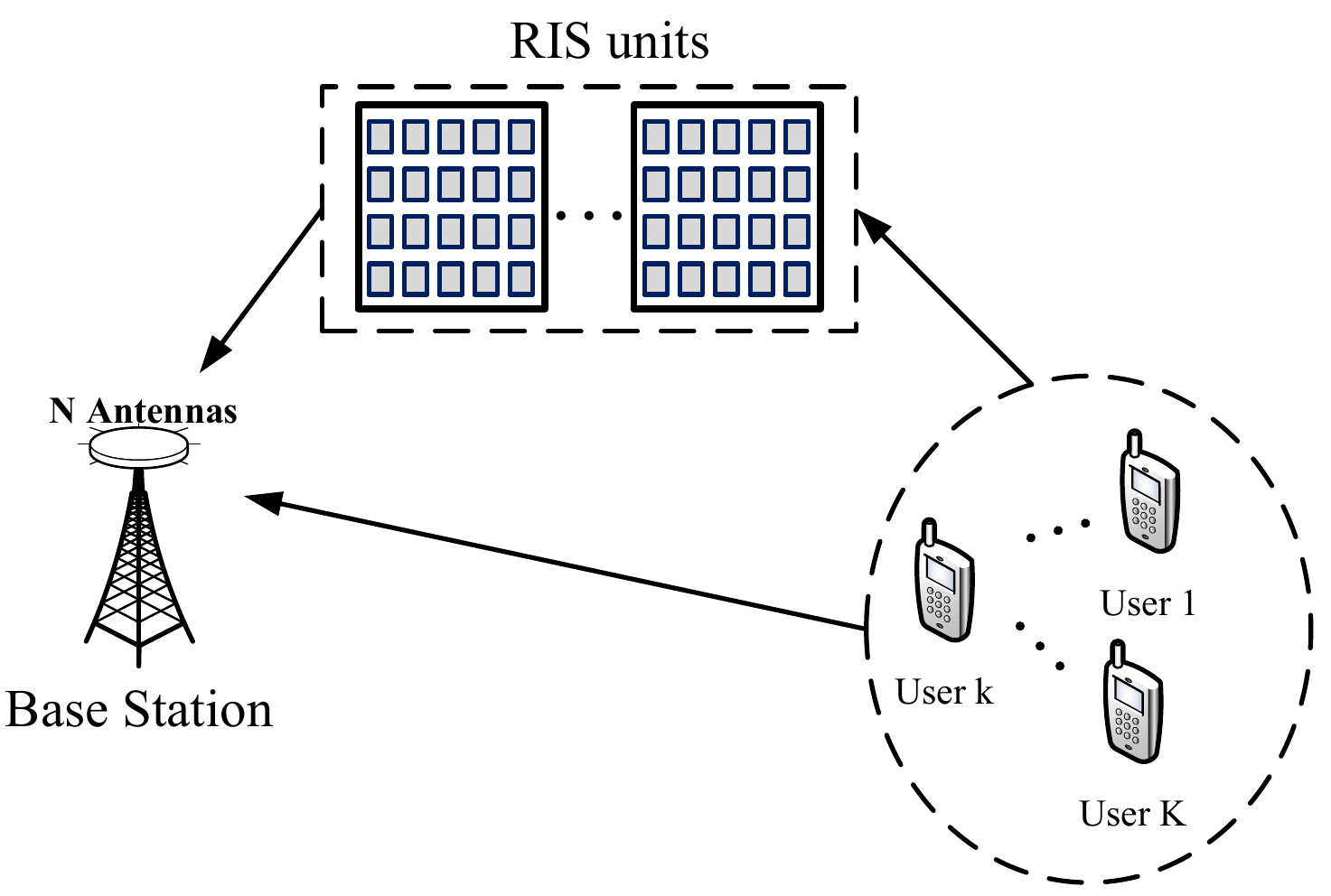}} \hspace{0.0in} 
	\subfigure[]{ 
		\label{fig:bachcost_user3}
		\includegraphics[width=3.1in]{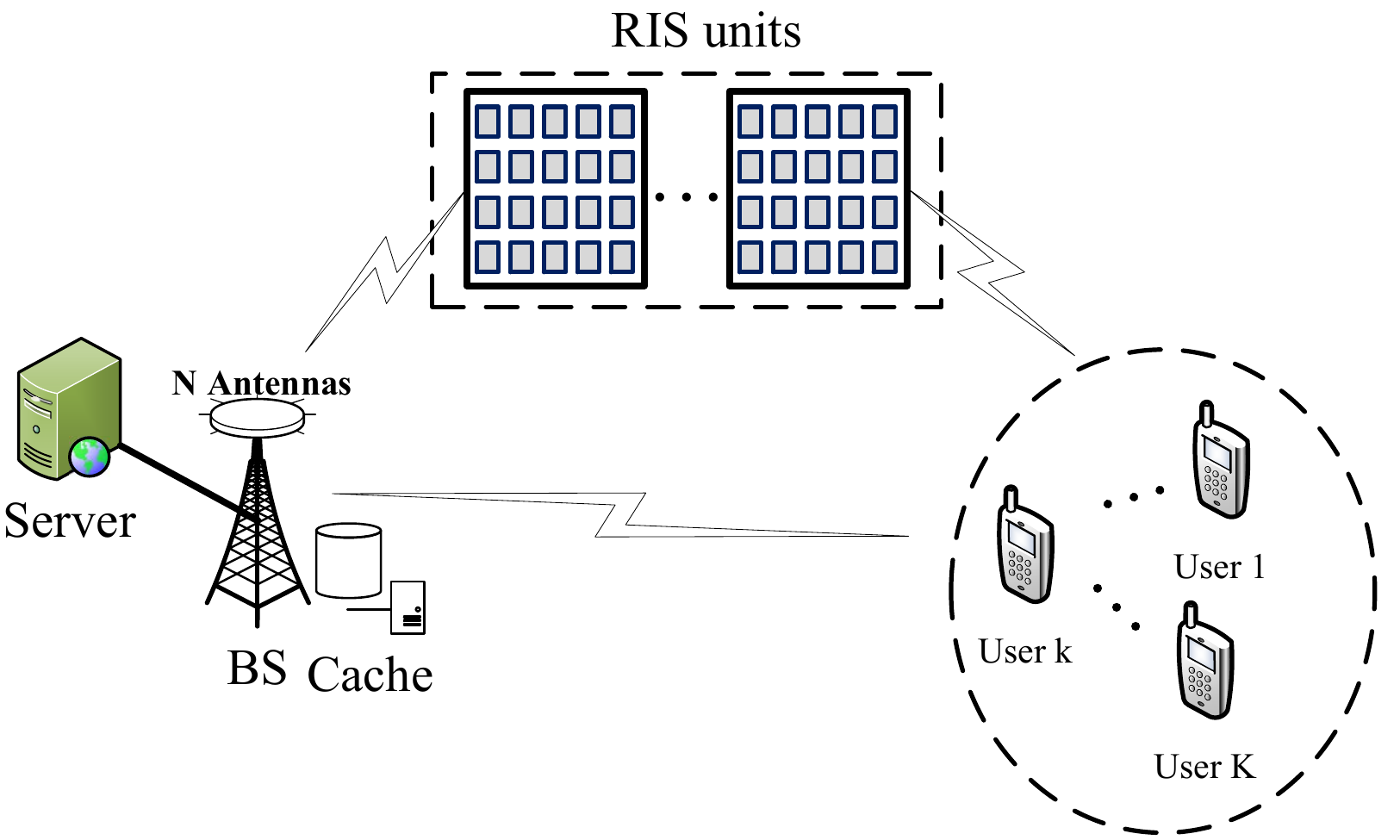}} \hspace{0.0in} 
	\caption{(a)~Secure beamforming for MISO systems.~(b)~MISO uplink communication networks.~(c)~Computation offloading in the IoT networks.}
\end{figure*} 
\begin{itemize}
\item Secure beamforming for multiple-input single-output (MISO) systems~\cite{J_LU20_RISsecure}: As shown in Fig.~\ref{fig:Se_EE_Power1}, the BS communicates with a single-antenna user with the help of an RIS in the presence of a single-antenna eavesdropper. The goal is to maximize the achievable secrecy rate by jointly optimizing the beamformer at the BS and the phase shift coefficients of the RIS under the transmit power constraint at the BS. To be specific, let the channels from the BS to the RIS, from the RIS to user, from the RIS to eavesdropper, and the beamforming vector at the BS be respectively denoted by $\mathbf{H}\in \mathbb{C}^{M\times N}$, $\mathbf{h}\in \mathbb{C}^{M\times 1}$, $\mathbf{g}\in \mathbb{C}^{M\times 1}$, and $\mathbf{w}\in \mathbb{C}^{N\times 1}$. Then, the secrecy rate maximization problem is given by
 \begin{subequations}\label{opt:example_1}
 	\begin{align}
\max\limits_{\mathbf{w},\mathbf{e}}~~& \log _{2}\left(\frac{\sigma^{2}+\left|\mathbf{e}^{H} \operatorname{diag}\left(\mathbf{h}^{H}\right) \mathbf{H} \mathbf{w}\right|^{2}}{\sigma^{2}+\left|\mathbf{e}^{H} \operatorname{diag}\left(\mathbf{g}^{H}\right) \mathbf{H}\mathbf{w}\right|^{2}}\right),\\
 \mathrm{s.t.}~~ &\|\mathbf{w}\|^2\leq P_\mathrm{max},	\\
 &\mathbf{e}\in \mathcal{F},
 	\end{align}
 \end{subequations}
where $\sigma^2$ is the variance of the white Gaussian noise at the user.
\item MISO uplink communication networks~\cite{J_LiuY20_RISImperfect}: 
There are a number of single-antenna mobile users transmitting signals to a multi-antenna BS with the assistance of an RIS, as shown in Fig.~\ref{fig:UPower_user2}. The objective is to minimize the total uplink transmit power by jointly optimizing the phase shift coefficients of the RIS $\mathbf{e}$, the transmission power ${x}_k$ of the user $k$ under the limited transmission power $P_k$, and signal-to-interference-and-noise-ratio (SINR) constraints. Let the channels from the BS to the RIS, from the RIS to user $k$, and from the BS to user $k$ be respectively denoted by $\mathbf{H}\in \mathbb{C}^{M\times N}$, $\mathbf{h}_{r,k}\in \mathbb{C}^{M\times 1}$, $\mathbf{h}_{d,k}\in\mathbb{C}^{N\times 1}$ with $k \in\{1,\ldots, K\}$. Accordingly, the weighted power minimization problem is given by
\begin{subequations}\label{opt:example_2}
	\begin{align}
\min\limits_{\mathbf{x},\mathbf{e}}~&\boldsymbol{\lambda}^T\mathbf{x},\\
\mathrm{s.t.}~&{x}_k\leq P_k,~\forall k,\\
&\mathbf{e}\in \mathcal{F},\\
&x_k\hat{\mathbf{h}}_k\left(\sigma^2\mathbf{I}_N+\sum_{i\neq k}x_i\hat{\mathbf{h}}^H_i\hat{\mathbf{h}}_i\right)^{-1}\hat{\mathbf{h}}_k^H\geq r_k,~\forall k
	\end{align}
\end{subequations}
where $\hat{\mathbf{h}}_k=\mathbf{h}_{r,k}^H\mathrm{diag}(\mathbf{e})\mathbf{H}+\mathbf{h}^H_{d,k}\in \mathbb{C}^{1\times N}$ is the equivalent channel from user $k$ to the BS, $\boldsymbol{\lambda}=[\lambda_1,\ldots,\lambda_K]^T$ represents the weights for mobile users, and $r_k$ is the minimum SINR requested by the user $k$.

\item Computation offloading in the Internet of Things (IoT) networks~\cite{J_chen20exploiting}: 
In the downlink transmission of an RIS-aided cache-enabled radio access network, a multi-antenna BS transmits signals to a number of single-antenna users, as shown in Fig.~\ref{fig:bachcost_user3}. The goal is to minimize the total network cost that consists of both the backhaul capacity and the transmission power by adjusting the caching proportion of the file requested by user~$k$, the precoding vector $\mathbf{p}_k \in \mathbb{C}^{M\times 1}$ at the BS for user $k$, and the RIS coefficients. Besides the constraint on the RIS coefficients, we also have a constraint on the size of total cached content to be smaller than the local storage size $S_{\max}$ at the BS. Further let the target rate of user $k$ be denoted by $R_k$, the total network cost minimization problem is formulated as
\begin{subequations}\label{opt:exaple3_min}
	\begin{align}
	\min\limits_{\mathbf{x},\mathbf{e},\{\mathbf{p}_k\}_{k=1}^K}~ 
	&\sum^K_{k=1}(1-x_k)R_k+\eta\sum^K_{k=1}\|\mathbf{p}_k\|^2,   \\
	\mathrm{s.t.}\quad
	&\sum_{k=1}^K {x}_k\leq S_{\max},\\
	& {x}_k\in [0,1],~\forall k\\
	& \mathbf{e}\in \mathcal{F},\\
	&\frac{|\hat{\mathbf{h}}_k\mathbf{p}_k|^2}{\sum_{l\neq k}|\hat{\mathbf{h}}_k\mathbf{p}_l|^2+\sigma^2}\geq 2^{R_k/B}-1, ~\forall k.
	\end{align}
\end{subequations}
where $\eta$ is a regularization parameter, $2^{R_k/B }-1$ is the SINR
requirement in terms of the content-delivery target rate of
user $k$, $B$ is the bandwidth of the system, and $\hat{\mathbf{h}}_k$ is defined as in the previous example.  
\end{itemize}

In the above three applications, we can see that most of the constraints in the resource allocation problems are decoupled in the sense that constraints for the RIS coefficents $\mathbf{e}$ does not involve other resources, and vice versa. For the coupled constraints, e.g., (2d) and (3e), they can be converted into penalty terms in the objective function \cite{Wupena, J_ly_CRAN} or decoupled by introducing auxiliary variables \cite{J_ZZ20SecureProb,WSTCOM, J_ly_eeprecoding, J_ly_cache}. After these operations, without loss of generality, we consider a general resource allocation problem appearing in the form
\begin{equation}\label{opt:oorig_reAll_RIS}
\min\limits_{\mathbf{x},\mathbf{e}}~~f\left(\mathbf{x},\mathbf{e}\right),\quad \mathrm{s.t.}~~\mathbf{x}\in \mathcal{X},~\mathbf{e}\in \mathcal{F},
\end{equation}
where $f\left(\mathbf{x},\mathbf{e}\right)$ is a continuous objective function, and $\mathbf{x}$ represents the conventional communication resources with the set $\mathcal{X}$ representing the constraint on $\mathbf{x}$, such as maximum transmit power, limited cache
size, operation time limitation, etc.

With the decoupled constraints for $\mathbf{x}$ and $\mathbf{e}$, the
optimization problem is tractable under the commonly used block cooridinate descent (BCD) framework,
which alternatively solves for $\mathbf{x}$ with $\mathbf{e}$ fixed and solves for $\mathbf{e}$ with $\mathbf{x}$ fixed. In particular, when the phase shift coefficients of the RIS $\mathbf{e}$ is given, the resource allocation problem reduces to a traditional communication problem without the RIS, which has been investigated for decades and should be familiar to many communication researchers.
On the other hand, when $\mathbf{x}$ is fixed at a certain value, say $\mathbf{x}^{(n)}$, the subproblem for optimizing $\mathbf{e}$ is  
\begin{equation}\label{opt:modulus_theta}
	\min_{\mathbf{e}}~~ 
	f\left(\mathbf{x}^{(n)},\mathbf{e}\right), 
	~~\mathrm{s.t.}~~
	\mathbf{e}\in \mathcal{F}.
\end{equation}

Before discussing various methods for solving $\eqref{opt:modulus_theta}$, let us review the modeling of the constraint set $\mathcal{F}$ on the RIS coefficients. Depending on whether the phase is modeled as a continuous or discrete variable, the feasible set $\mathcal{F}$ is defined differently:
\begin{itemize}
	\item  $\mathbf{Continuous~phase~shift}$: Each RIS coefficient has infinite phase resolution, i.e., $e_m$ is expressed as $\beta_m e^{i\theta_m}$ with $i$ being the imaginary unit, and $\theta_m$ is a real number. For $\beta_m$, there are three variations in the literature. 
	
	\begin{itemize}
	    \item C1. $\beta_m$ is a known constant, which is the ideal phase shift model \cite{wuc1,huc1,huac1}. This is the most popular model at the time of writing, and $\mathcal{F}$ is represented by modulus constraints $|{e}_m|^2 = 1$;
	    \item C2. $\beta_m$ is an unknown variable and is independent of $\theta_m$ \cite{yangc2,zhaoc2}. This model leads to a convex set $\mathcal{F}$, described by $|{e}_m|^2 \leq c$ for some constant $c$;
	    \item C3. $\beta_m$ is a function of $\theta_m$. This is a relatively new model and takes the hardware property into consideration. For example, one of the recent models \cite{J_Abeywickrama20IdealRIS} states that 
\begin{equation}
\beta_{m}\left(\theta_{m}\right)=\left(1-\beta_{\min }\right)\left(\frac{\sin \left(\theta_{m}-\phi\right)+1}{2}\right)^{\alpha}+\beta_{\min }, \nonumber
\end{equation}
where $\beta_{\min}$, $\phi$ and $\alpha$ are known constants related to the specific circuit implementation.
\end{itemize}
	\item $\mathbf{Discrete~phase~shift}$: Each RIS coefficient $e_m$ can only take one of the $L$ possible phase shift values.
\end{itemize}

Among the three continuous phase shift models, C2 is a convex set, thus its treatment is similar to the conventional resource allocation problem. Another way to view C2 is by treating the optimization of $\beta_m$ and $\theta_m$ separately, so C2 is equivalent to $0 \leq \beta_m \leq \sqrt{c}$ and $|e^{i\theta_m}|^2=1$. If we regard optimization of $\beta_m$ as part of conventional resources, the remaining constraint $|e^{i\theta_m}|^2=1$ reduces to model C1. For C3, although it is non-convex, it can be handled by gradient descent method on $\theta_m$ (to be detailed in the next section). For C1, even though $\beta_m$ is known and fixed, due to the modulus requirement, its handling is non-trivial and there exists a number of methods  with different solution qualities for tackling this constraint. 

On the other hand, for the discrete phase shift case, the corresponding problem~\eqref{opt:modulus_theta} is an integer nonlinear program and is NP-hard (i.e., the optimal solution cannot be found in polynomial time). However, the most prevalent way for handling this model is to relax the discrete variables to their continuous counterparts. Then each of the obtained continuous phase shifts (by any methods for solving continuous phase shift model) is quantized to its nearest discrete value. Since the resolution of discrete phase shifts increases with the number of allowable phases, the quantization loss will be insignificant when the number of allowable phases is large \cite{WuDIS}. 

Since C1 is the most fundamental model, in this paper, we focus on reviewing the optimization methods for model C1, with some of the reviewed methods also applicable to C2 and C3. The AI-based methods will be covered in Section~\ref{Sec:V_learning} with reinforcement learning also suitable for discrete phase shift model. Further emerging approaches for handling discrete phase shift case will be discussed in the section of future challenges.

\section{Review on Optimization Methods under Continuous Phase Shift}
\label{Sec:ContinousRIS}
Currently, the major techniques for optimizing the continuous phase shifts are the semi-definite relaxation (SDR) method, penalty method, majorization minimization (MM) method, graident descent (GD) method, manifold method, and convex relaxation (CR) method. All the reviewed methods are primarily developed for C1, and can be applied to C2 if $\beta_m$ and $\theta_m$ are optimized separately. For C3, it is handled by GD method due to the complicated dependence of $\beta_m$ on $\theta_m$. Table I provides a quick summary of the reviewed methods in this section.

\newcommand{\tabincell}[2]{\begin{tabular}{@{}#1@{}}#2\end{tabular}}  
\begin{table}[!htbp]
	\centering
	\caption{Comparison of optimization methods for continuous  phase shift RIS}\label{tab:OptimizationMethodTable}
	\begin{tabular}{@{}ccccc@{}}
		\toprule
		Optimization Methods & Property of Solutions & Complexity Order & Applicable Model & Examples Using This Method  \\
		\midrule
		SDR & infeasible/feasible solution &  $\mathcal{O}\left(\sqrt{M}(2M^4+M^3)\right)$ & C1 and C2 & \cite{J_Qing19IRS_beamform}\cite{J_YouChannelEst20}\\
		\midrule
		Penalty  & stationary solution & $\mathcal{O}(M^3)$ & C1 and C2 &\cite{J_XuRISSecure20}\cite{TSP_pen} \\
		\midrule
		MM & locally optimal solution~\cite{J_Sun17MM}&  $\mathcal{O}\left(M^2\right)$ & C1 and C2 & \cite{J_Huang19RIS_EE} \cite{J_Wang20SecureRIS_CSI}\\
		\midrule
		GD & stationary solution~\cite{J_ma20lowcomplexity} & $\mathcal{O}\left(M\right)$ & C1, C2, and C3 &\cite{J_Nadeem20RIS_sinr} \cite{J_liu20beamformImpairment}\\
		\midrule
		Manifold & stationary solution~\cite{B_AbsilP09} & $\mathcal{O}\left(M\right)$ & C1 and C2 & \cite{J_chen20exploiting}\cite{letter_mani}\\
		\midrule
		CR & feasible solution~\cite{J_ma20lowcomplexity} &   \tabincell{c} {$\mathcal{O}\left(M^3\right)$ using CVX \\$\mathcal{O}\left(M\right)$ using PG} & C1 and C2 & 	\cite{J_Chen19RISSecure}\cite{J_guo2019weighted} \\
		\bottomrule
	\end{tabular}
\end{table}

\subsection{SDR Method}
To handle the nonconvex modulus constraints, we can introduce a rank-one auxiliary variable $\mathbf{Q}=\mathbf{e}\mathbf{e}^H$. This translates the optimization variable from $\mathbf{e}$ to $\mathbf{Q}$, and the objective function changes from $f\left(\mathbf{x}^{(n)},\mathbf{e}\right)$ to $f(\mathbf{x}^{(n)},\mathbf{Q})$. To account for the rank-one property of $\mathbf{Q}$ and the diagonal elements of $\mathbf{Q}$ are all 1, we need to add constraints $\mathrm{rank}(\mathbf{Q})=1$ and $\mathbf{Q}_{m,m}=1, \forall m$.
 Then, problem~\eqref{opt:modulus_theta} under C1 is equivalent to
\begin{subequations}\label{opt_FP_theta_Q}
	\begin{align}
	\min_{{\mathbf{Q}\succeq \mathbf{0}}}\quad 
	&f\left(\mathbf{x}^{(n)},\mathbf{Q}\right),  \label{obj:FP_Dinkel_theta} \\
	\mathrm{s.t.}\quad
	&\mathbf{Q}_{m,m}=1,~~\forall m,\\
	& \mathrm{rank}(\mathbf{Q})= 1.  \label{cons:rank_1_Dinkel}
	\end{align}
\end{subequations}
Notice that the transformed problem is still intractable due to the rank constraint $\mathrm{rank}(\mathbf{Q})=1$. But the celebrated SDR method (i.e., removing the rank constraint) can be employed to solve this problem if the cost function $f\left(\mathbf{x}^{(n)},\mathbf{Q}\right)$ is convex in $\mathbf{Q}$. 

More specifically, with the remaining constraints $\mathbf{Q}_{m,m}=1$ for $m=1,\ldots,M$ being transformed into semidefinite constraints $\operatorname{Tr}\left(\mathbf{E}_{m} \mathbf{Q}\right)=1$ for $m=1, \ldots, M$, where $\left\{\mathbf{E}_{m} \in \mathbb{C}^{M \times M}\right\}_{m=1}^{M}$ is a matrix with a single 1
in the $(m, m)^{t h}$ position and zero in all other positions, variable $\mathbf{Q}$ can be directly updated via the interior point method, which is available in the popular software package CVX. If $f$ is not convex, we may add another layer of successive convex approximation (SCA) to convexify the objective function in each SCA iteration with the complexity increased by a factor equal to the number of iterations for SCA.
However, since the rank-1 constraint is relaxed, the obtained solution may not be a feasible solution to the original problem (6).

In general, a feasibility check is used to verify whether the obtained $\mathbf{Q}$ satisfies the rank constraint. Since the relaxed problem is a convex problem, a closed-form solution of $\mathbf{Q}$ or explicit expression with respect to $\mathbf{Q}$ can be derived in its dual domain. Then the feasibility check can be done by leveraging the ranks product inequalities technique \cite{Liletter}.
If the rank constraint is not satisfied, a Gaussian randomization procedure can be employed to extract a feasible solution~\cite{J_LuoSDP10}. 
Since the computational complexity order of SDR is $\mathcal{O}\left(\sqrt{M}(2M^4+M^3)\right)$, it could be too time-consuming for large-scale RISs.

\subsection{Penalty Method}
To guarantee a feasible solution while avoiding the feasibility check of the SDR method, a penalty method can be employed.
To be specific, the rank constraint $ \mathrm{rank}(\mathbf{Q})= 1$ in \eqref{cons:rank_1_Dinkel} can be equivalently expressed as 
$\mathrm{Tr}(\sqrt{\mathbf{Q}^*\mathbf{Q}})-\|\mathbf{Q}\|_2\leq 0$~\cite{B_NumOpt06}, where $\mathbf{Q}^*$ is the conjugate of $\mathbf{Q}$. Then, with the constraint added as a penalized term, this further transforms problem (6) into
\begin{subequations}\label{opt:penalty_problem_DC}
	\begin{align}
	\min_{{\mathbf{Q}\succeq \mathbf{0}}}\quad 
	&f\left(\mathbf{x}^{(n)},\mathbf{Q}\right) + \frac{1}{\mu}\left(\mathrm{Tr}(\sqrt{\mathbf{Q}^*\mathbf{Q}})-\|\mathbf{Q}\|_2\right), \label{obj:Penalty_Q_initial} \\
	\mathrm{s.t.}\quad
	&\mathbf{Q}_{m,m}=1,~~\forall m, \label{cons:trace_penalty_DC}
	\end{align}
\end{subequations}
where $\mu\in (0,1)$ is a penalty factor penalizing the violation of constraint $\mathrm{Tr}(\sqrt{\mathbf{Q}^*\mathbf{Q}})-\|\mathbf{Q}\|_2\leq 0$. This transformed objective function now contains a difference-of-convex (DC) term
$\mathrm{Tr}(\sqrt{\mathbf{Q}^*\mathbf{Q}})$ $-\|\mathbf{Q}\|_2$.
To convert the DC term to a convex form, SCA can be applied to $-\|\mathbf{Q}\|_2$ (if $f$ is non-convex, the SCA can also be applied to $f$ at the same time). 

Such a resultant problem is convex in $\mathbf{Q}$ if $f\left(\mathbf{x}^{(n)},\mathbf{Q}\right)$ is convex. Accordingly, the optimal $\mathbf{Q}$ in each SCA iteration can be obtained by employing the interior-point method.
Since the transformed problem is solved under the SCA framework, a stationary solution of $\mathbf{Q}$ can be guaranteed. Furthermore, since problem~\eqref{opt:modulus_theta} is equivalent to the transformed problem as $\mu$ tends to zero, the obtained solution is also a stationary point to~\eqref{opt:modulus_theta}.
The penalty factor $\mu$ is important in controlling how strict the rank constraint is imposed. In practice, it can be a decreasing sequence with respect to the SCA iteration to guarantee a feasible solution of~\eqref{opt:modulus_theta} at the end of the iteration.
As the interior-point method is adopted in each SCA iteration, the complexity order is at least $\mathcal{O}(M^3)$.

\subsection{MM Method}
\begin{figure*}	
	\centering
	\subfigure[]{
		\label{alg:MM_convex_linear} 
		\includegraphics[width=3.1in]{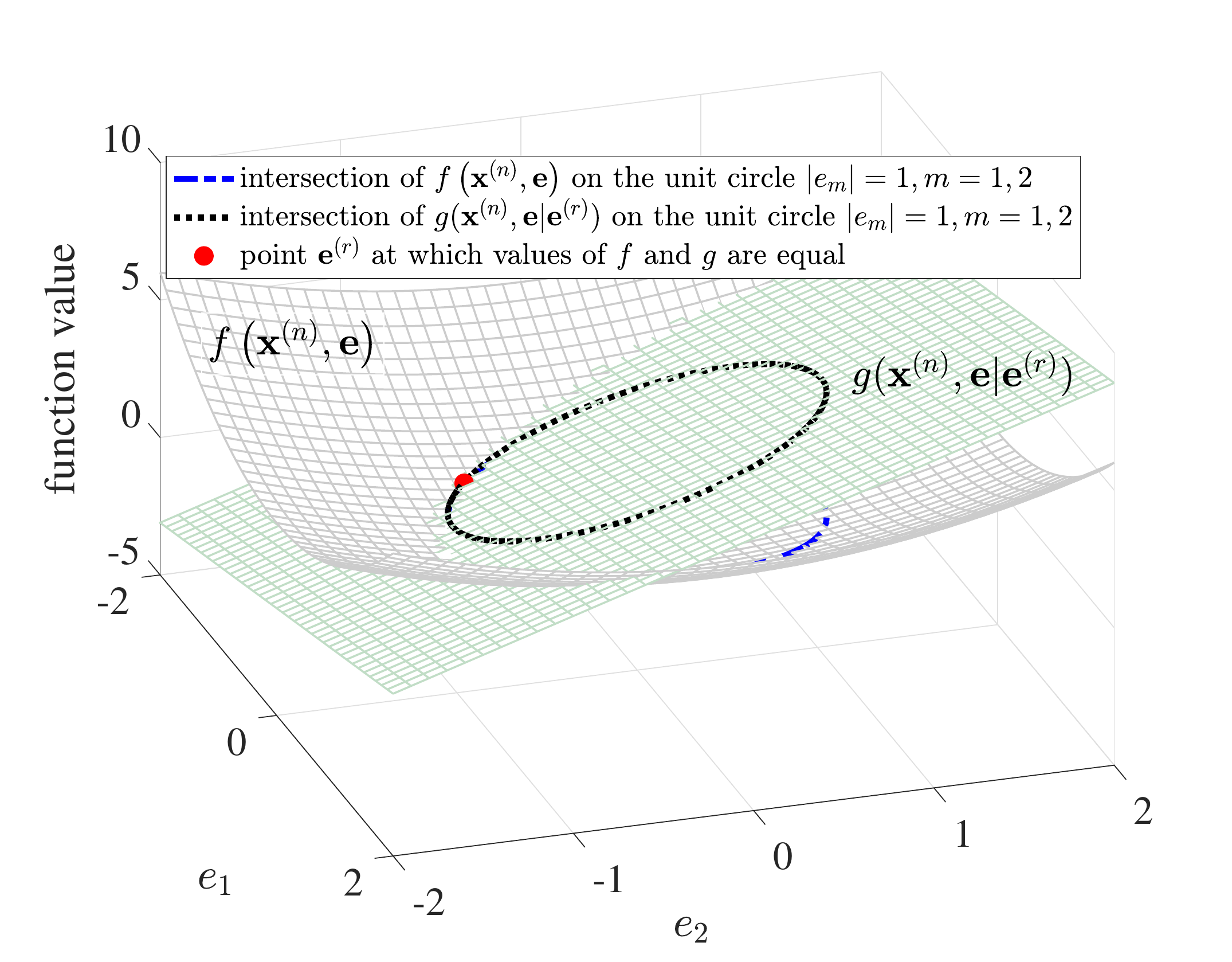}} \hspace{0.0in} 
	\subfigure[]{
		\label{alg:GD_demo} 
		\includegraphics[width=3.1in]{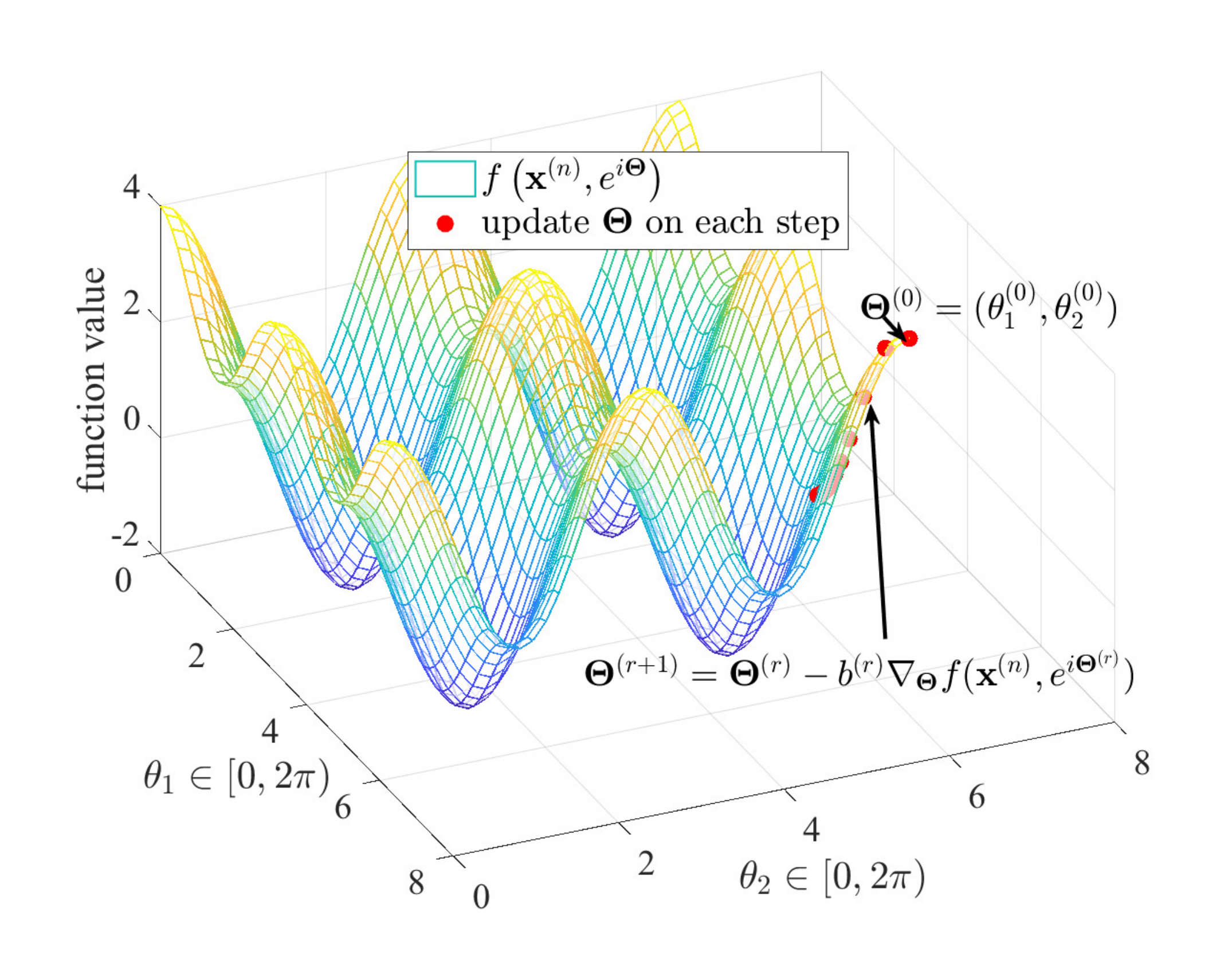}} \hspace{0.0in} 
	\caption{(a)~Linear upperbound $g(\mathbf{x}^{(n)},\mathbf{e}|\mathbf{e}^{(r)})$ for a quadratic function $f(\mathbf{x}^{(n)},\mathbf{e})$ at point $\mathbf{e}^{(r)}$ on the unit circle.~(b)~Graphical representation of the GD method for updating $\mathbf{\Theta}$, where $b^{(r)}$ is the update step size and $\nabla_{\mathbf{\Theta}} f\left(\mathbf{x}^{(n)}, e^{i\mathbf{\Theta}^{(r)}}\right)$ is the gradient of $f$ at the last iteration solution $\mathbf{\Theta}^{(r)}$.} 
\end{figure*}

Both the SDR method and the penalty method require a complexity at least $\mathcal{O}(M^3)$. To reduce the computational complexity, the MM method can be employed to tackle the unit-modulus constraint. The key idea lies in constructing a sequence of surrogate functions that serve as upper bounds of the cost function with respect to the unknown variable $\mathbf{e}$. Figure~\ref{alg:MM_convex_linear} visualizes how a linear surrogate function $g(\mathbf{x}^{(n)},\mathbf{e}|\mathbf{e}^{(r)})$ upper bounds a convex quadratic function $f\left(\mathbf{x}^{(n)},\mathbf{e}\right)$ on the unit circle at the $r^{th}$ iteration. 
Specifically, given the solution of $\mathbf{e}$ at the $r^{th}$ iteration as $\mathbf{e}^{(r)}$ (the red point in Fig.~\ref{alg:MM_convex_linear}), the constructed linear surrogate function needs to satisfy: a) $g(\mathbf{x}^{(n)},\mathbf{e}|\mathbf{e}^{(r)}) \geq f\left(\mathbf{x}^{(n)},\mathbf{e}\right)$ on the unit circle manifold;
b) $g(\mathbf{x}^{(n)},\mathbf{e}|\mathbf{e}^{(r)})=f\left(\mathbf{x}^{(n)},\mathbf{e}\right)$ at $\mathbf{e}^{(r)}$; c)
$\nabla_{\mathbf{e}} f\left(\mathbf{x}^{(n)},\mathbf{e}\right) = \nabla_{\mathbf{e}} g(\mathbf{x}^{(n)},\mathbf{e}|\mathbf{e}^{(r)})$ at point $\mathbf{e}^{(r)}$.
In practice, the second-order Taylor expansion and Jensen's inequality are commonly used to find $g(\mathbf{x}^{(n)},\mathbf{e}|\mathbf{e}^{(r)})$~\cite{J_Sun17MM}. 

With the established upper bound $g(\mathbf{x}^{(n)},\mathbf{e}|\mathbf{e}^{(r)})$, problem~\eqref{opt:modulus_theta} under C1 can be iteratively solved with subproblem at the $(r+1)^{th}$ iteration being 
\begin{equation}\label{opt:MM_g-modulus}
	\min_{\mathbf{e}}~~
	g(\mathbf{x}^{(n)},\mathbf{e}|\mathbf{e}^{(r)}), \\
\quad	\mathrm{s.t.}~~ |{e}_m|^2 = 1, \quad  \forall m. 
\end{equation}
Since $g(\mathbf{x}^{(n)},\mathbf{e}|\mathbf{e}^{(r)})$ is a linear surrogate function, it has a closed-form minimizer $\mathbf{q}_{\mathbf{e}^{(r)}}$. Then we can project $\mathbf{q}_{\mathbf{e}^{(r)}}$ onto the unit circle manifold to obtain $\mathbf{e}^{(r+1)}$. The next iteration involves finding $\mathbf{q}_{\mathbf{e}^{(r+1)}}$ based on $\mathbf{e}^{(r+1)}$, and the process repeats. Therefore, problem~\eqref{opt:modulus_theta} can be iteratively solved and the final converged point is a local optimal point of problem~\eqref{opt:modulus_theta}~\cite{J_Sun17MM}. 
The computational complexity of the MM method is dominated by the determination of surrogate functions, which gives a complexity order of $\mathcal{O}(M^2)$.

\subsection{GD Method}
Even with the MM method, the complexity order is quadratic.
To further reduce the computational complexity to linear order, GD can be employed to find a stationary point of~\eqref{opt:modulus_theta}. The key observation is that the ultimate unknown variable in the feasible set $\mathcal{F}$ is in fact $\{\theta_m\}_{m=1}^M$ instead of $\mathbf{e}$. Therefore, problem~\eqref{opt:modulus_theta} can be recast into an unconstrained optimization problem as  
\begin{equation}\label{eq:theta_phase_RA}
\min\limits_{\mathbf{\Theta}}~~f\left(\mathbf{x}^{(n)}, e^{i\mathbf{\Theta}}\right),~~\mathrm{s.t.}~~
\mathbf{\Theta}=[ \theta_1,\ldots,  \theta_M]^T.
\end{equation}
By recasting the quadratic function $f\left(\mathbf{x}^{(n)},\mathbf{e}\right)$ shown in Fig.~\ref{alg:MM_convex_linear} as $f\left(\mathbf{x}^{(n)}, e^{i\mathbf{\Theta}}\right)$, a graphic demonstration of the GD method is illustrated in Fig.~\ref{alg:GD_demo}. 
Using a feasible initialization point $\mathbf{\Theta}^{(0)}$, $\mathbf{\Theta}^{(r+1)}$ can be obtained at the $(r+1)^{th}$ iteration based on $\mathbf{\Theta}^{(r+1)} = \mathbf{\Theta}^{(r)} - b^{(r)}\nabla_{\mathbf{\Theta}} f\left(\mathbf{x}^{(n)}, e^{i\mathbf{\Theta}^{(r)}}\right)$, where $b^{(r)}$ is the step size. Since only gradient information is involved in each update, GD has a linear complexity order with respect to $M$, and the final converged point is a stationary solution to~\eqref{opt:modulus_theta}. 
Another point to note is that by expressing the objective function in terms of $\mathbf{\Theta}$, many local minimums are introduced compared to the objective function in terms of $\mathbf{e}$. Therefore, the quality of the converged solution of the GD method highly depends on the initialization. Notice that since this method directly optimize with respect to $\theta_m$, it is also applicable to model C3 where $\beta_m$ is a function of $\theta_m$. The only change in \eqref{eq:theta_phase_RA} is replacing $e^{i\mathbf{\Theta}}$ with $[\beta_1(\theta_1)e^{i\theta_1},\ldots,\beta_M(\theta_M)e^{i\theta_M}]^T$.

\subsection{Manifold Method}
\begin{figure}[tb]
	\centering
	\includegraphics[scale=0.65]{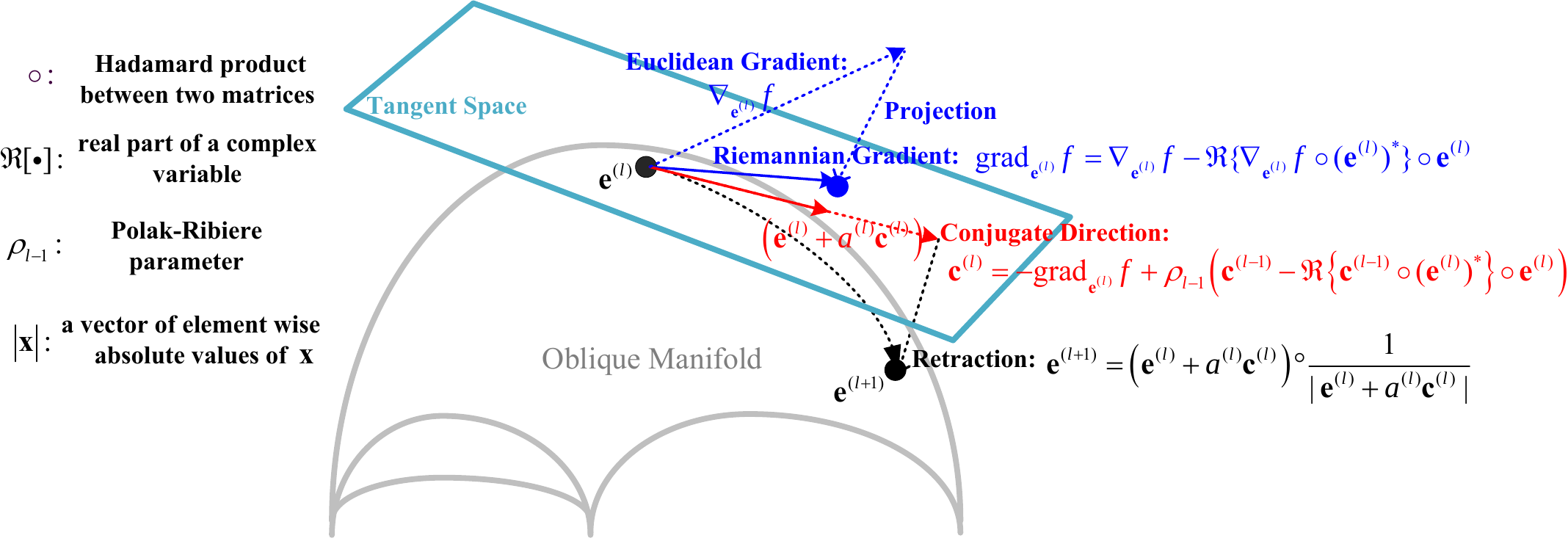}
	\caption{Graphic illustration of Riemannian CG method at the $l^{th}$ iteration.}\label{alg:CG_obligueManifold} 
\end{figure}

Recognizing the constraint
set $\mathcal{F}$ forms a complex circle manifold in model C1, another low-complexity method is based on manifold optimization. A representative algorithm in this category is the Riemannian conjugate gradient (CG) method~\cite{B_AbsilP09}, which solves problem~\eqref{opt:modulus_theta} on an oblique manifold through alternatively computing the Riemannian gradient, finding the conjugate direction, and performing retraction mapping. A graphical representation of various steps of the Riemannian CG method is illustrated in Fig.~\ref{alg:CG_obligueManifold}. More specifically, the Riemannian gradient of $f\left(\mathbf{x}^{(n)},\mathbf{e}\right)$ at the $l^{th}$ iteration solution $\mathbf{e}^{(l)}$ is obtained by projecting the Euclidean gradient of $f$ at $\mathbf{e}^{(l)}$ onto the tangent space (blue color step in Fig. 4).
After obtaining Riemannian gradient $\mathrm{grad}_{\mathbf{e}^{(l)}} f$, the CG descent direction at point $\mathbf{e}^{(l)}$ can be obtained as $\mathbf{c}^{(l)}$, and $\mathbf{e}^{(l)}$ is updated as $\mathbf{e}^{(l)}+{a}^{(l)}\mathbf{c}^{(l)}$ on tangent space, where ${a}^{(l)}$ is a Armijo backtracking step size (red color step in Fig. 4). 
Since the updated $\mathbf{e}^{(l)}+{a}^{(l)}\mathbf{c}^{(l)}$ may not be in the oblique manifold, the final point should be projected to the oblique manifold by employing retraction mapping (black color step in Fig.~4). 

This method extends the GD method in the Euclidean space to the Riemannian manifold. Compared to the GD method in the previous subsection, the manifold method does not re-formulate the objective function in terms of $\mathbf{\Theta}$ and thus avoids the many local minimums as shown in Fig.~\ref{alg:GD_demo}. By guaranteeing the complex circle constraint satisfied in every iteration, the Riemannian CG method converges to a stationary solution~\cite{B_AbsilP09}. The computational complexity of the Riemannian CG update is dominated by the gradient step which only involves element-wise operations. This gives a linear complexity order with respect to $M$. 

\subsection{CR Method}
The idea of the CR method is that while the constraint set $\mathcal{F}$ in C1 is nonconvex, it can be relaxed to a Euclidean unit ball, which is a convex set. Therefore, problem~\eqref{opt:modulus_theta} under C1 can be relaxed into
\begin{equation}
\min\limits_{\mathbf{e}}~~f\left(\mathbf{x}^{(n)}, \mathbf{e}\right),~~\mathrm{s.t.}~~
|{e}_m|^2 \leq 1, \quad  \forall m.
\end{equation} 
Since (10) has a convex set, it can be solved via convex tools such as CVX. 
Afterward, the solution of the relaxed problem is projected to the nearest point in $|e_m|^2=1$ to obtain a feasible solution.

A variant of the above method is replacing the interior point method with the projected-gradient (PG) method, which alternates between gradient step and projection step. Although this variant has not been employed in the existing literature involving RISs, it has a linear computational complexity compared to the cubic complexity of the interior point method, thus is promising for large-scale systems. 

Notice that this method is applicable to model C2. For model C2, where $\mathcal{F}$ is already in the form of $|e_m|^2 \leq c$, there is no relaxation involved and the solution is directly obtained from solving (10).  Furthermore, unlike other methods applying to C2, there is no need to optimize $\beta_m$ and $\theta_m$ separately since optimization of $\beta_m$ is incorporated in (10).  

\subsection{Summary and Performance Comparison}
\begin{figure*}	
	\centering
	\includegraphics[scale=0.55]{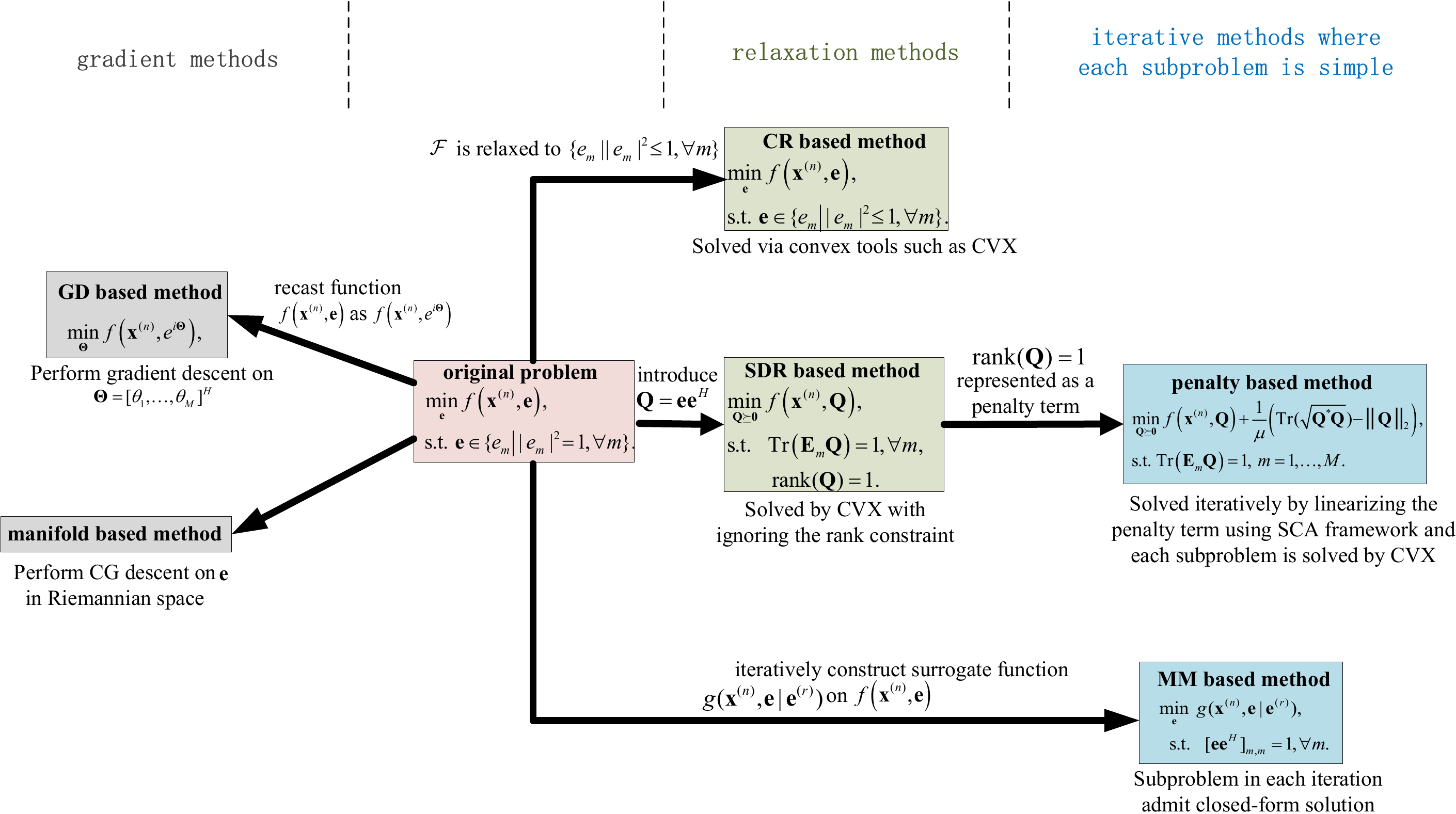}
	\caption{Relationships among different optimization methods.}\label{alg:Relationship_demo}
\end{figure*}
To sum up, the optimization methods for handling continuous phase shift design in this section can be categorized into relaxation methods (SDR and CR), iterative approximation methods (penalty-based method and MM), and gradient methods (GD and manifold method). Their relationships are summarized in Fig.~\ref{alg:Relationship_demo} and their properties are compared in Table~\ref{tab:OptimizationMethodTable}.

\begin{figure*}	
	\centering
	\subfigure[]{ 
		\label{fig:Se_EE_Power}
		\includegraphics[width=3.1in]{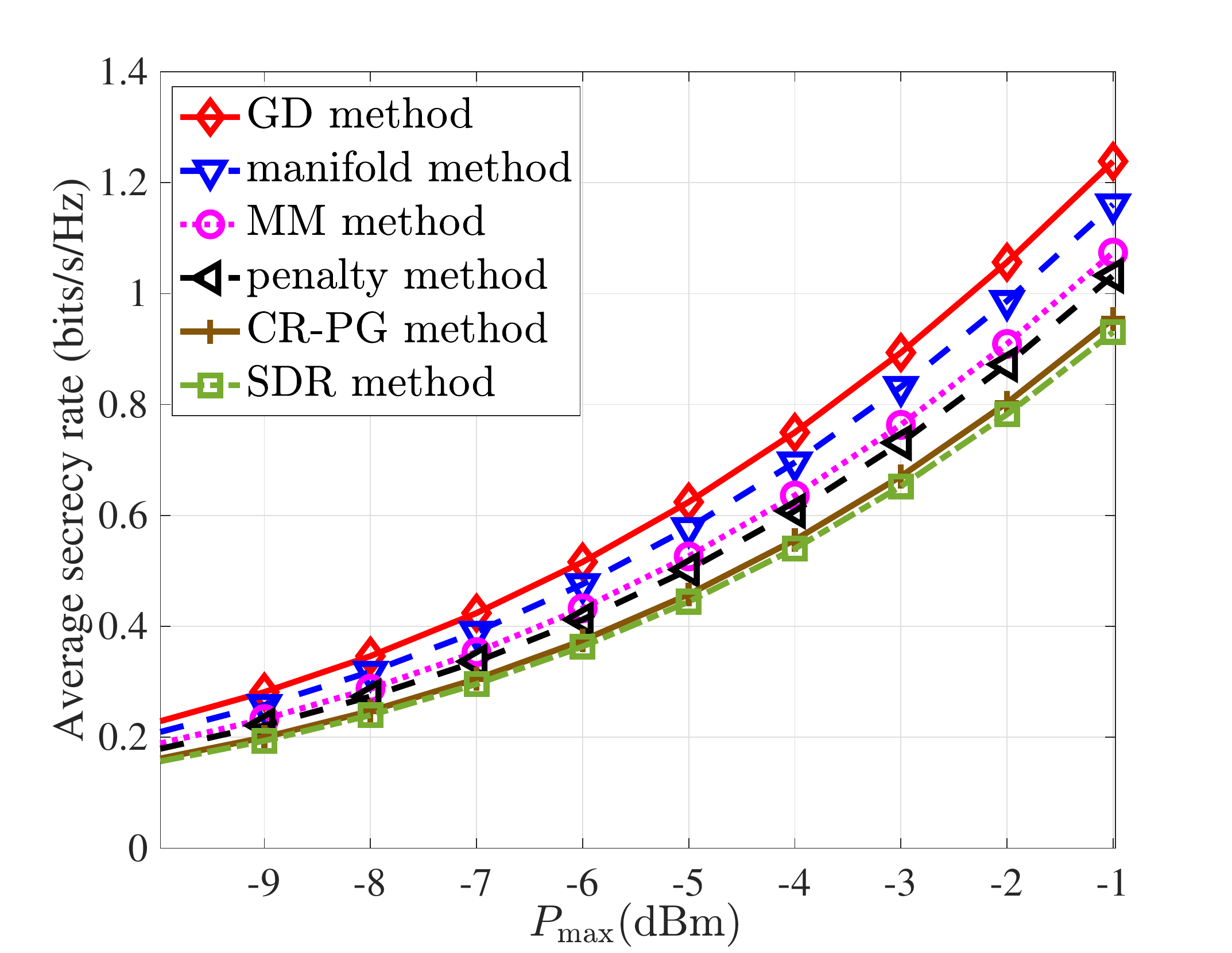}} \hspace{0.0in} 
	\subfigure[]{ 
		\label{fig:UPower_user}
		\includegraphics[width=3.1in]{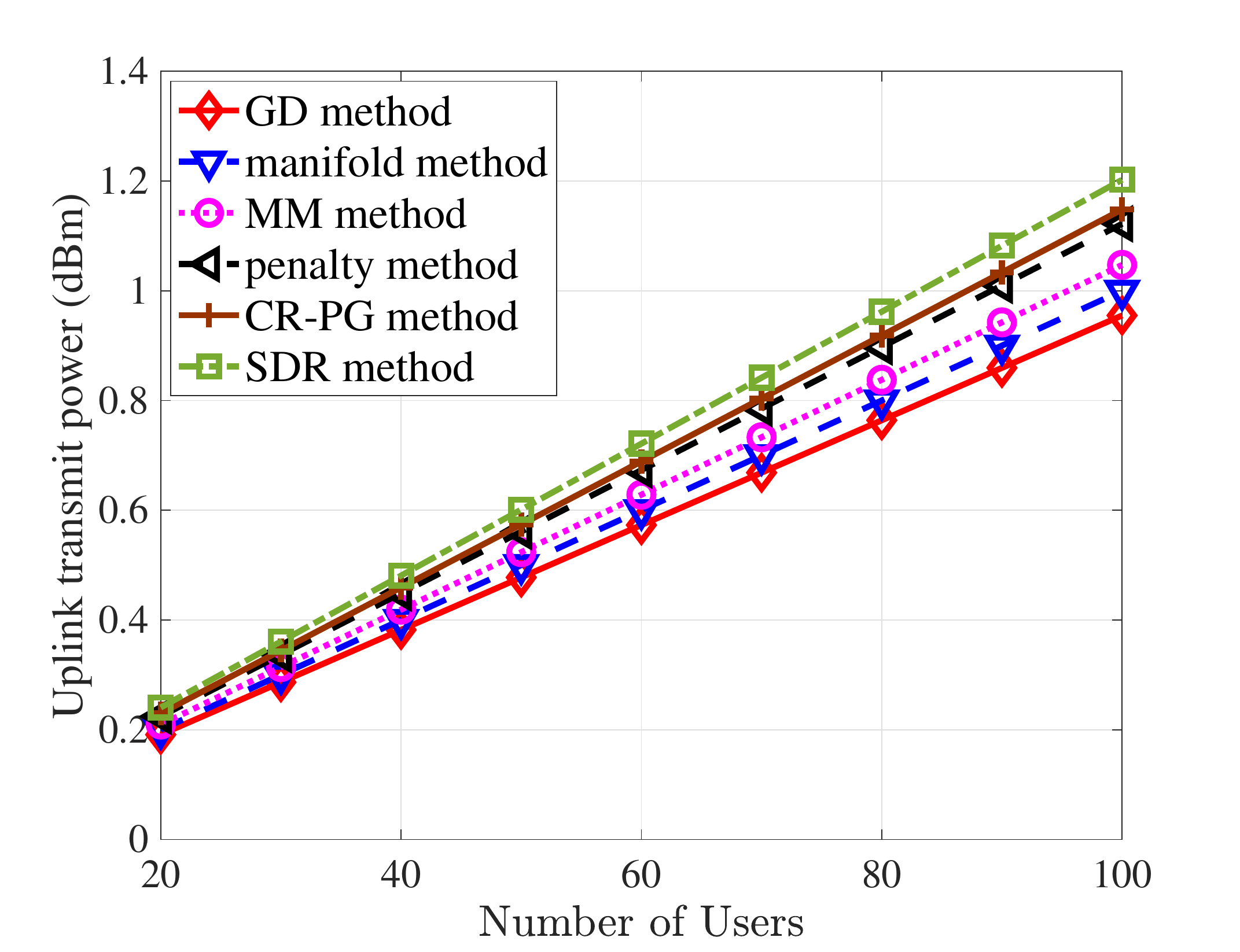}} \hspace{0.0in} 
	\subfigure[]{ 
		\label{fig:bachcost_user}
		\includegraphics[width=3.1in]{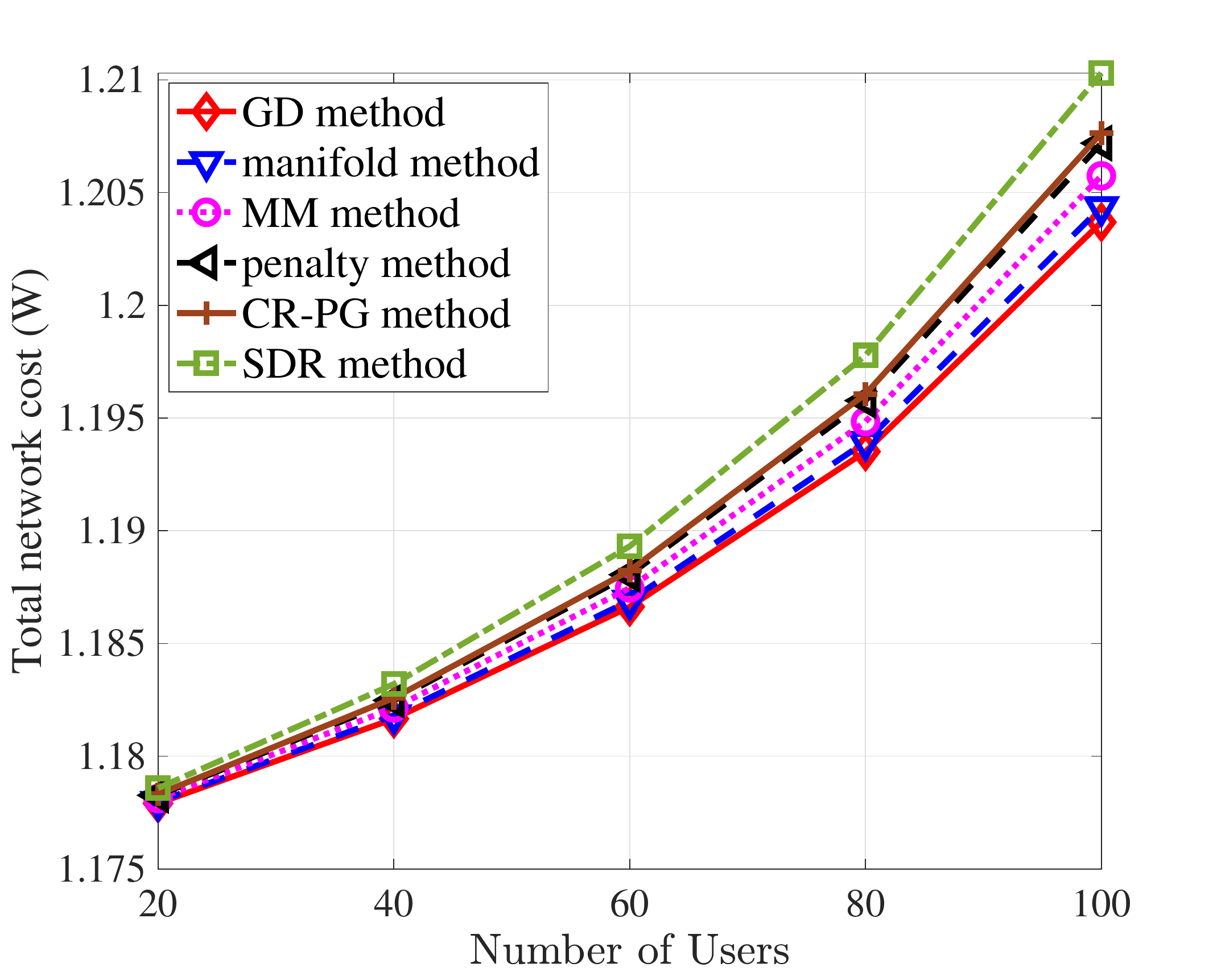}} \hspace{0.0in} 
	\caption{Performance comparisons of six optimization methods with $M=10$: (a)~Secrecy rate versus the maximum transmit power under the number of BS antennas $N = 20$~\cite{J_LU20_RISsecure}.~(b)~Uplink transmit power versus the number of users under the number of BS antennas $N = 20$ and transmission power limitation $P_k = 10.8$~dBm~\cite{J_LiuY20_RISImperfect}.~(c)~Total network cost versus the number of users under the number of BS antennas $N = 10$, the target rate $R_k = 10$~MHz, the bandwidth $B=10$~MHz, regularization parameter $\eta=100$ and local storage size $S_{\max}=100$~\cite{J_chen20exploiting}.}
\end{figure*}

To compare the performance of different optimization methods, the three application examples mentioned in Section~\ref{Sec:II} are simulated under phase shift model C1. All simulations are performed on MATLAB R2017a on a Windows X64 desktop with 3.2 GHz CPU and 16 GB RAM. For fair comparisons, all algorithms start from the same initial point (any feasible point can serve as an initial point) and the stopping criterion for iterative methods is when the relative change of two consecutive objective function values being less than $10^{-4}$, and the maximum number of iterations for all methods is set to $100$.
By employing the BCD framework for solving $\mathbf{x}$ and $\mathbf{e}$ in~(4), the three applications can be efficiently solved and the simulation results are shown in Figs.~\ref{fig:Se_EE_Power},~\ref{fig:UPower_user}, and~\ref{fig:bachcost_user}, respectively. 

From these figures, it can be observed that out of the six algorithms, GD and the manifold method perform consistently well in all three applications, followed by the MM method and the penalty method. On the other hand, the SDR method and CR perform the worst in these three applications. The worse performance of the SDR method and CR is due to the relatively weak guarantee in the solution quality.

\begin{figure*}	
	\centering
	\subfigure[]{
		\label{fig:Se_EE_time} 
		\includegraphics[width=3.1in]{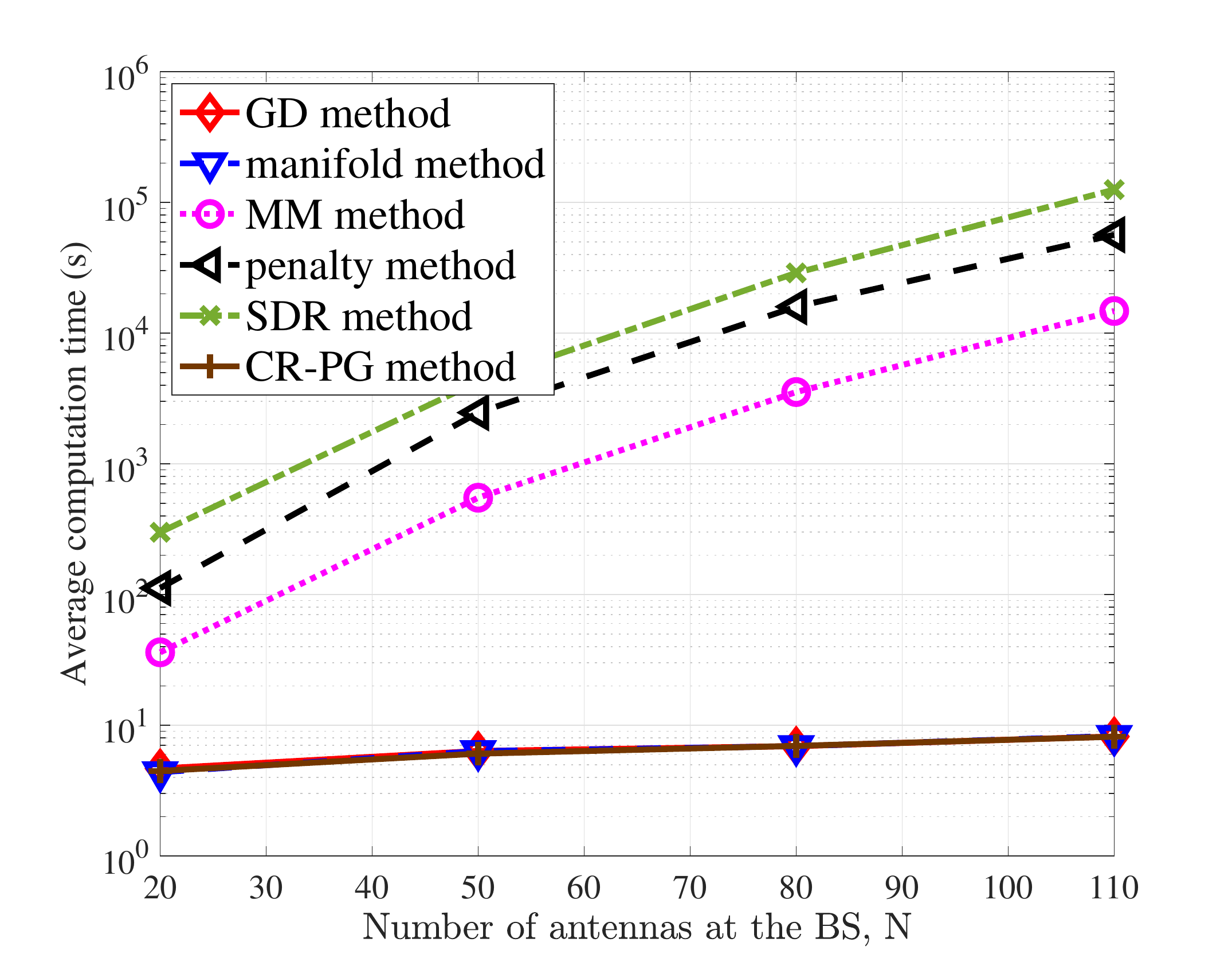}} \hspace{0.0in} 
	\subfigure[]{
		\label{fig:reflecting_EE_time} 
		\includegraphics[width=3.1in]{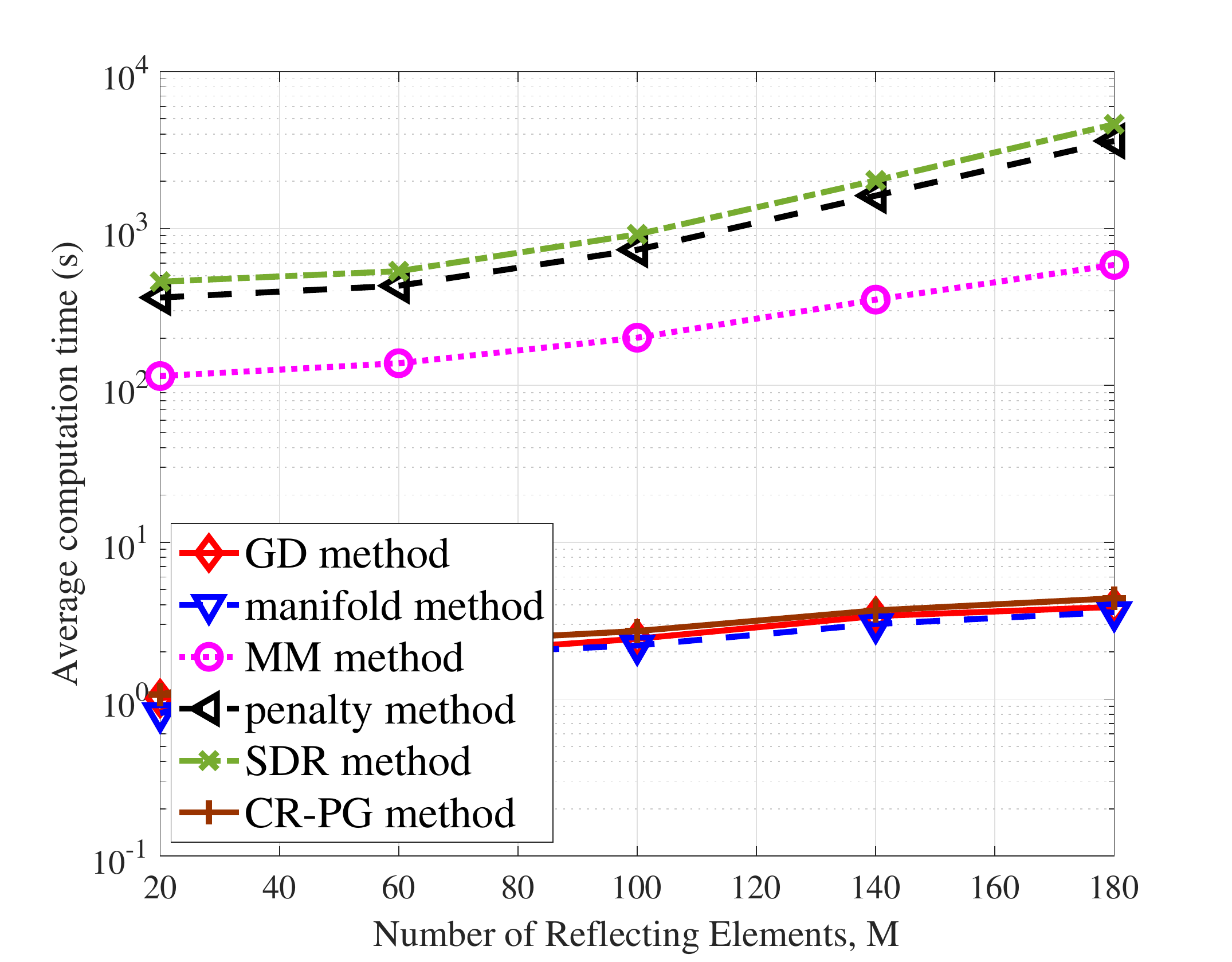}} \hspace{0.0in} 
	\caption{Performance comparisons of six optimization methods: (a)~Average computation time versus the number of antennas at the BS when $M=10$.~(b)~Average computation time versus the number of reflecting elements when $N=10$.} \label{fig:performTime_otherBaLiScheme}
\end{figure*}

On the other hand, the computation times of various methods in the first application are shown in Fig.~\ref{fig:Se_EE_time}. From this figure, it can be seen that the manifold method, the GD method, and the CR-PG method require the least amount of computation times among the six algorithms, achieving at least two orders of magnitude reduction compared with the SDR method and the penalty method when $N>50$. This advantage becomes more prominent as the number of reflecting elements $M$ increases as shown in Fig.~\ref{fig:reflecting_EE_time}. The computation times for other two applications show similar behaviors and thus are not shown here.

\section{Learning to Optimize An RIS}\label{Sec:V_learning}
In addition to mathematical optimization methods, AI-based methods have recently emerged as a promising direction for solving
resource allocation problems. Problem~\eqref{opt:modulus_theta} can be regarded as a regression problem (or classification problem for discrete phase shifts), which can be tackled by deep learning (DL) methods. When DL is employed, a deep neural network (DNN) is adopted to learn the mapping from the channel state information (CSI) to the optimized phase shift coefficients. Once the AI model is trained, the computation of phase shift coefficents is extremely fast, and it can be readily implemented in various operating systems such as Linux and Android via model loading. In the following, three learning-based methods are discussed.

\subsection{Supervised Learning}	
In this paradigm, the optimal phase shift $\mathbf{e}$ under a specific channel realization and network setting is obtained by traditional optimization approaches (as detailed in Section~\ref{Sec:ContinousRIS}). This channel realization and the corresponding optimized phase shift are treated as a training sample. If we have many training samples corresponding to different channel realizations, a DNN can be trained to approximate the behavior of a traditional optimization method. The advantage of this approach is that the learning results inherit the solution quality from optimization methods~\cite{J_Hu21LearningOptimize_RIS,C_Taha19DL_RIS}. However, it has an additional burden of generating training samples, although low-complexity methods such as GD, the manifold method and CR-PG help to reduce this burden compared to SDR and MM methods. Furthermore, supervised learning can be extended to directly solve problem~\eqref{opt:oorig_reAll_RIS} by treating the channel realization as input and all resources (both $\mathbf{x}$ and $\mathbf{e}$) as the desired output of the DNN.

\subsection{Unsupervised Learning}
The connection between unsupervised learning and problem~\eqref{opt:modulus_theta} comes from the observation that~\eqref{opt:modulus_theta} can be regarded as an unconstrained optimization problem if the variable is viewed in terms of $\theta_m$ instead of $\mathbf{e}$. This view has been adopted in the GD method in Section III. But in contrast to the GD method for solving~\eqref{eq:theta_phase_RA} with respect to $\mathbf{\Theta}$, unsupervised learning uses a DNN that accepts channel realization as input and generates the corresponding $\mathbf{\Theta}$ as output, where the optimization is with respect to the coefficients of the DNN. In unsupervised learning, the objectve is to minimize $\mathbb{E}[f(\mathbf{x}, e^{i\mathbf{\Theta}})]$, where the expectation is with respect to the distribution of input channel state information. The training procedure involves first generating a large number of channel realizations, then optimizing $\mathbf{\Theta}$ and $\mathbf{x}$ under the BCD framework. When optimizing $\mathbf{\Theta}$, back propagation is used. On the other hand, when optimizing $\mathbf{x}$, conventional optimization technique is used with the expectation tackled via sampling approximation. Different from supervised learning, this approach does not need the labelling of data, which saves a significant amount of time in training data preparation. However, a disadvantage is that the obtained solution does not have any quality guarantee.
	
\subsection{Reinforcement Learning }
Another major framework in AI is the deep reinforcement learning (DRL). In this framework, the agent (i.e., decision maker) gradually derives its best action through trial-and-error interactions with the environment over time. There are a few basic elements characterizing the DRL learning process: the state, the action, the reward, and the state action value function.
\begin{enumerate}
	\item $\mathbf{State}$: a set, denoted by $S$, characterizing the environment. The state $s^{(t)} \in S$ denotes the environment at the time step $t$.
	\item $\mathbf{Action}$: a set of allowable action, denoted by $A$. Once the agent takes an action $a^{(t)} \in A$ at time instant $t$ (determined by the state action value function), the state of the environment will transit from the current state $s^{(t)}$ to the next state $s^{(t+1)}$. 
	\item $\mathbf{Reward}$: the performance metric of a particular action, denoted by $r^{(t)}$ at time instant $t$.
	\item $\mathbf{State~action~value~ function~(Q-function)}$:~while the reward represents immediate return from action $a$ at state $s$, the state action value function indicates cumulative rewards the agent may get from taking action $a$ in the state $s$, which is denoted by $Q\left(s, a\right)$.
\end{enumerate}
Depending on the types of action spaces, two DRL methods are available:  the deep Q-network (DQN) algorithm, which is designed for discrete action spaces, and the deep deterministic policy gradient (DDPG), which is designed for continuous action spaces. Hence, DQN fits the discrete phase shift model, while the DDPG is employed for continuous phase shift variables. 

In this subsection, we present a mapping of DQN in the context of resource allocation problems in RIS empowered wireless networks. In this model, the central controller, which controls the RIS, acts as the agent. At each time slot $t$, the agent observes a state, $s^{(t)} \in S$, which consists of all channel state information from the wireless system. According to the current state and the $Q$-function, the agent takes an action, $a^{(t)} =\operatorname{argmax}_{a} Q\left(s^{(t)}, a\right) \in A$, where $A$ consists of discrete phase shifts that each reflecting element is allowed to take. After carrying out an action $a^{(t)}$, the agent obtains a reward $r^{(t)}$ determined from the negative objective function of~\eqref{opt:modulus_theta} and observes the next state $s^{(t+1)}$ generated by the wireless system. At each time slot, $Q\left(s^{(t)}, a^{(t)}\right)$ is updated by 
\begin{equation}
Q\left(s^{(t)}, a^{(t)}\right) = Q\left(s^{(t)}, a^{(t)}\right) + \alpha \left(r^{(t)} + \gamma \operatorname{argmax}_{a} Q\left(s^{(t+1)}, a\right) - Q\left(s^{(t)}, a^{(t)}\right) \right),\nonumber
\end{equation}
where $\alpha$ is the learning rate and $\gamma$ is the discount factor designed for DQN. The aim of the DQN model is to enable the agent to carry out actions to maximize the long-term sum reward.


\subsection{Summary and Performance Comparison}

\begin{figure*}	
	\centering
	\includegraphics[scale=0.28]{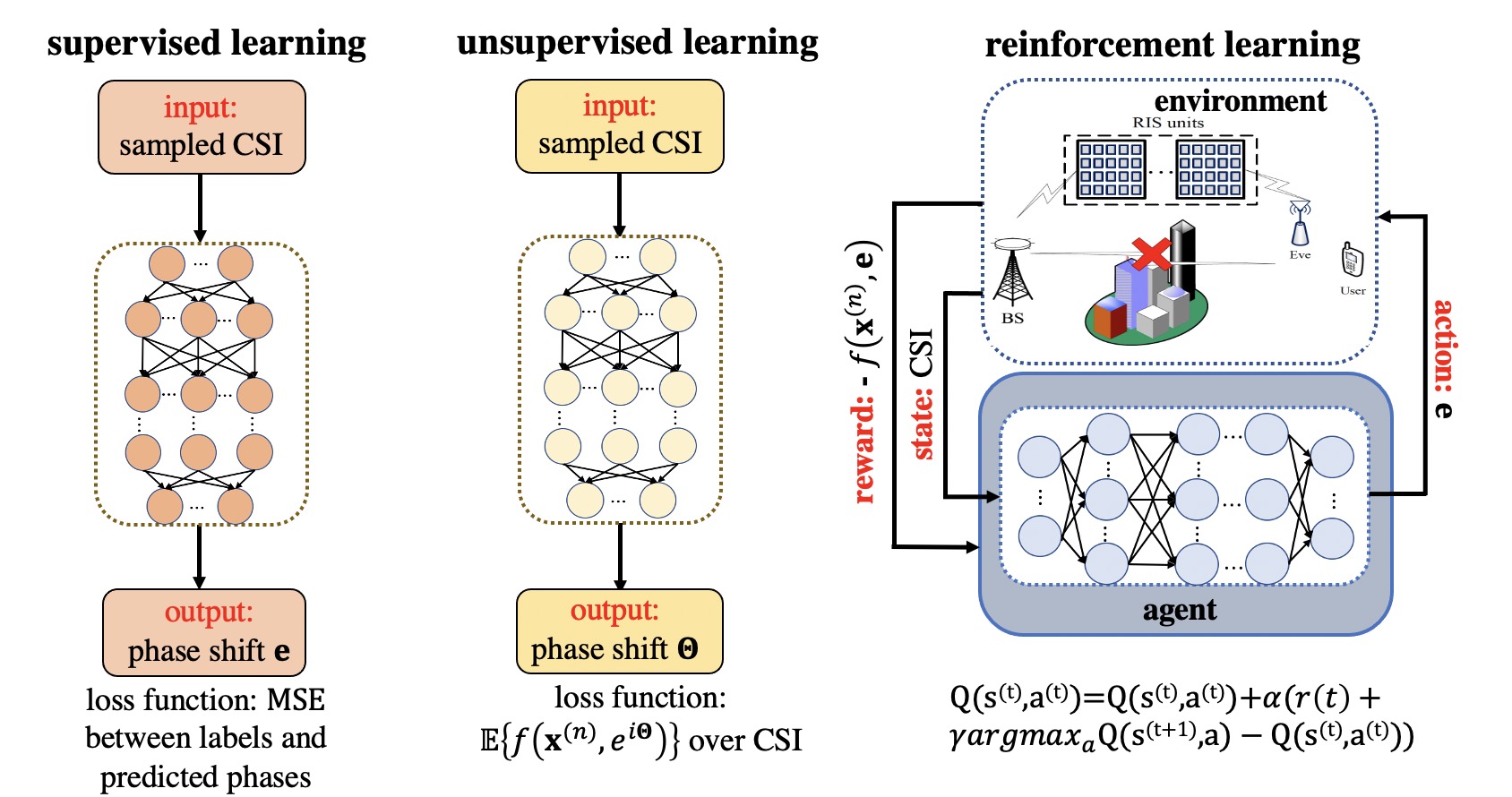}
	\caption{Illustration of different learning methods. The loss functions of supervised learning and unsupervised learning are the mean squared error (MSE) between labels and predicted phases, and the expectation of the objective function of~\eqref{eq:theta_phase_RA} over CSI, respectively.}	\label{alg:Relationship_learn}
\end{figure*}

Different learning-based methods for solving problem~\eqref{opt:modulus_theta} are summarized in Fig.~\ref{alg:Relationship_learn}. For supervised learning, since the training samples are generated from conventional optimization methods, the quality of the output is determined by the properties of the solution from the employed optimization method. For the other two methods (unsupervised learning and reinforcement learning), the outputs have no such quality guarantee. 

To compare the performance of different learning-based methods, the first example mentioned in Section~\ref{Sec:II} is simulated, with the GD optimization selected for generating training samples in supervised learning, and also serves as a performance benchmark. Fig.~9(a) shows the case of continuous phase shift. It is clear that supervised learning performs close to the GD algorithm. This is not surprising as supervised learning is mimicking the behavior of the optimization method chosen for generating the training data. However, for unsupervised learning, although it does not need training data preparation, it performs unmistakably worse than the supervised learning. Table~\ref{tab:LearningMethodTable} further shows the training times and inference times of GD, supervised learning, and unsupervised learning. It can be observed that the inference times of deep learning methods are indeed very short compared to GD method, although their preparation and training times are very long.

\begin{figure*}	
	\centering
	\subfigure[]{ 
		\label{fig:Pero_Learning}
		\includegraphics[width=3in]{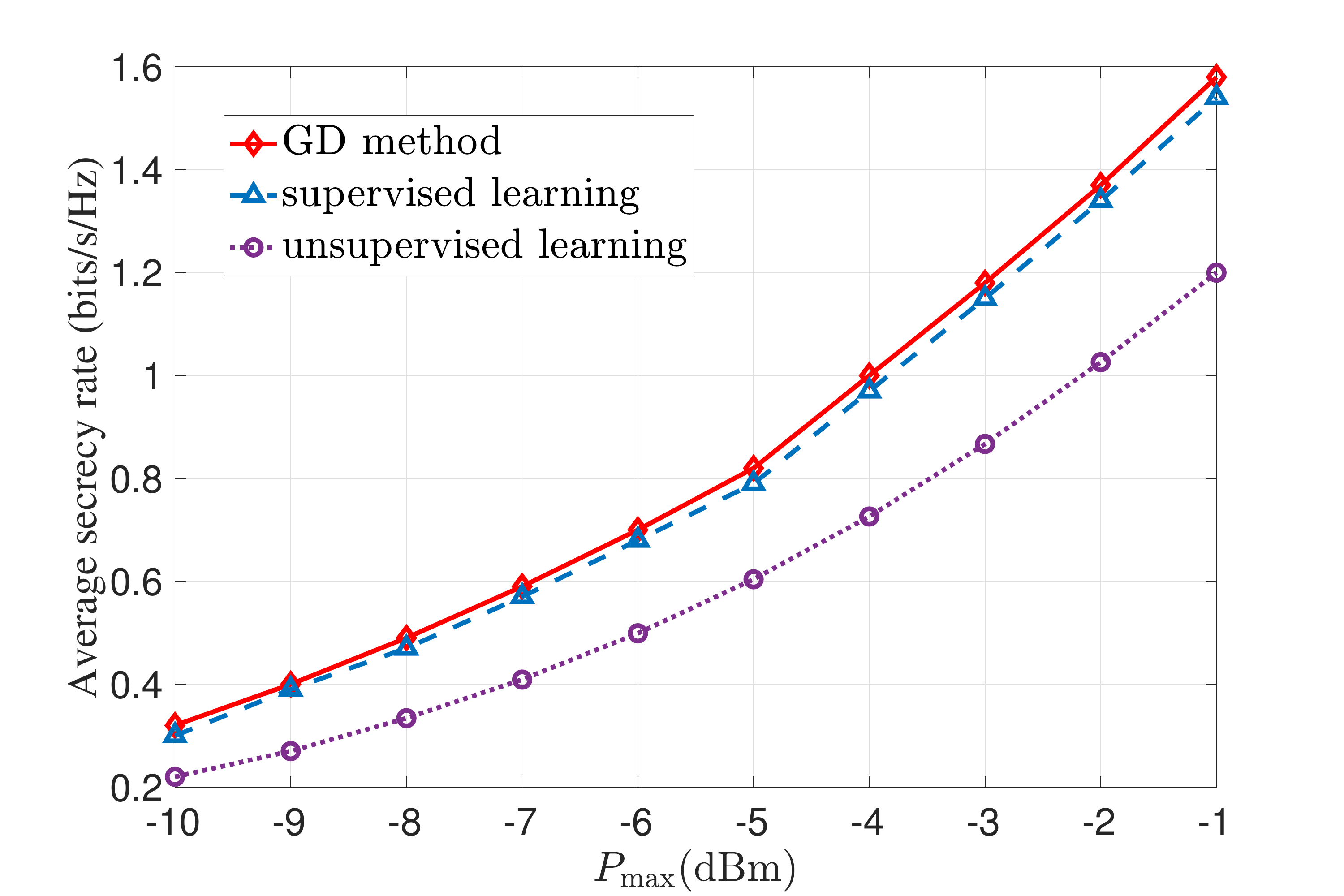}} \hspace{0.0in}
	\subfigure[]{ 
		\label{fig:perform_otherBaLiScheme}
		\includegraphics[width=3in]{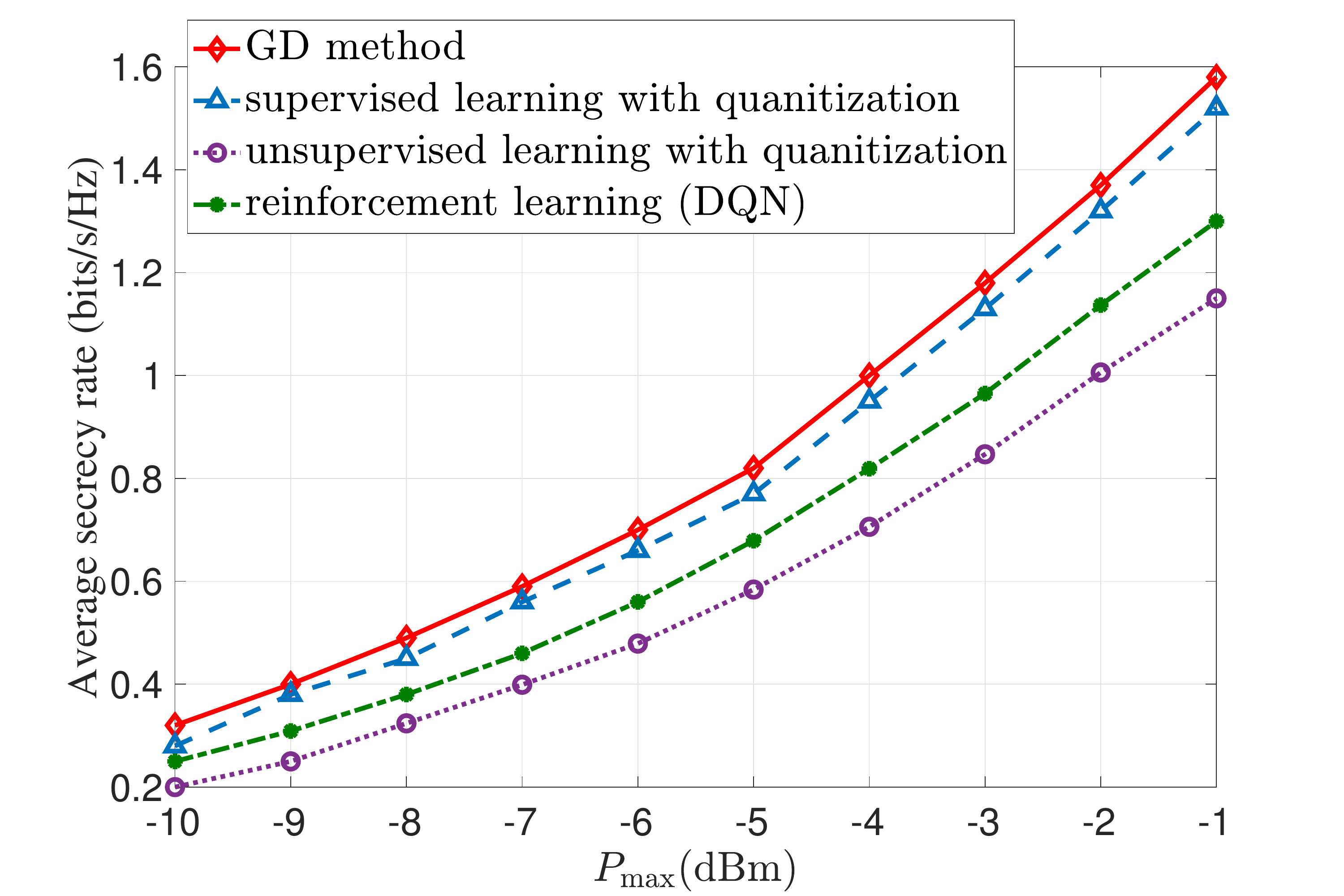}} \hspace{0.0in}
	\caption{Performance comparison of different learning-based methods in the secrecy rate optimization problem under $N=10$, $M=10$. All the simulations are implemented on Colab with TensorFlow 2 and backend GPU. For training, 800,000 independent channels are generated (with corresponding optimized phase obtained by GD methods for supervised learning).  For testing, 200,000 independent channels are used. 
 The adopted neural networks consist of 3 fully-connected hidden layers, containing 500, 250, 200 neurons, respectively. The rectified linear unit (ReLu) is used as the activation function for the hidden layers and the linear activation function is applied to the output layer. (a)~continuous phase case.~(b)~discrete phase case.}
\end{figure*}

On the other hand, the performance of deep learning methods under eight allowable discrete phases is shown in Fig.~\ref{fig:perform_otherBaLiScheme}. For the supervised learning and unsupervised learning, we simply apply quanitization to the learning results. For DRL, we employ the the DQN algorithm, which is trained with a DNN for 2000 epochs and 128 minibatches for each epoch. GD method with unquantized output is also included in Fig.~\ref{fig:perform_otherBaLiScheme} to show the performance limit. It can be seen from Fig.~\ref{fig:perform_otherBaLiScheme} that the performance of quantization under supervised and unsupervised learning do not degrade much compared to the unquantized output in Fig.~\ref{fig:Pero_Learning}. For DQN, its performance lies between supervised learning and unsupervised learning. The training and inference times of DQN are also shown in Table~\ref{tab:LearningMethodTable}.
\begin{table}[!htbp]
	\centering
	\caption{Comparison of learning methods for phase shift RIS}\label{tab:LearningMethodTable}
	\begin{tabular}{@{}cccc@{}}
		\toprule
		Methods & Training data preparation time & Training Time & Inference Time \\
		\midrule
		GD & N/A & N/A &  21.7 ms \\
	    \midrule
		Supervised Learning & 4.8 h& 10.521 h  &  87.1 $\mu$s\\
		\midrule
		Unsupervised Learning & N/A  & 11.347 h & 66.3 $\mu$s \\
		\midrule
		Reinforcement Learning & N/A & 17.862 h & 14.3 ms \\
		\bottomrule
	\end{tabular}
\end{table}

\section{Future Challenges}
While an explosive growth in the number of studies of resource allocation involving RISs has been witnessed in the past few years, there are still challenging problems remaining to be investigated. Below, four challenges are described and potential solutions are also discussed.

\subsection{Handling Channel Uncertainty}
In general, due to the large number of passive reflective elements in RISs, imperfect CSI is inevitable. Considering channel uncertainty, the resource optimization would be a stochastic counterpart of the problems discussed earlier. In particular, the CSI random error would make the constraints appear in a probabilistic form and the objective function takes an extra expectation. 

If the distribution of the channel uncertainty is known, this statistical information can be used to transform the probabilistic constraints into deterministic ones and compute the expectation of the objective function explicitly \cite{J_20NOMASecureGrad,yc_tsp,zzltwc}. However, due to the cascaded channel created by the RIS, the statistical information of the CSI might be complicated, making the transformation from stochastic problems into deterministic ones suffers performance loss, and/or expectation computation intractable. In that case, the Monte Carlo simulation-based method could be used to handle the channel uncertainty \cite{luedtke2008sample}.

On the other hand, learning-based methods can be modified to tackle uncertain CSI, even when the distribution of the channel uncertainty cannot be described in closed-form. In particular, when preparing the training data, we generate both the true CSI and the CSI added with uncertainty. During the training, we input the observed CSI (which contains errors) to the DNN, but compute the loss function or reward function using error-free CSI. In this way, the learning system can automatically learn to ``denoise'' the CSI, while learning the mapping of the RIS phase shifts.

\subsection{Handling Discrete Phase Shift}

Recently, the discrete phase shift model begins to emerge under the argument that the reflecting elements only have finite reflection levels due to hardware limitations. The resulting resource allocation problem is even more challenging than the continuous phase shift counterpart since the problem involves both continuous and discrete variables.  At the moment of writing, there are two major techniques for solving discrete phase shift problems: quantization or brute-force searching, with the majority of works adopting quantization.

For the quantization-based method, we have demonstrated in Fig. 9(b) that the performance loss is insignificant if the number of discrete phase shifts is not very small. This explains why quantization-based method is popular among existing works. However, when the number of allowable phases is small (e.g.,  2 or 3), the quantization method will lead to inevitable performance degradation. To overcome this issue, the original integer nonlinear program can be iteratively transformed into integer linear programs via linear cuts. Then, the branch-and-bound algorithm and exhaustive search can be employed to handle the resultant problem with discrete variables \cite{jsac_song}. However, these searching methods have an exponential time complexity, which could lead to unacceptable complexity even for modest values of $M$. Recently, the idea of alternating optimization (AO) has been applied to the discrete phase shift searching \cite{J_WuRIS_20Discrete}, in which multiple phase shifts are optimized one at a time so that the searching space in each iteration is small. While this reduces the complexity significantly, only stationary points can be guaranteed.

As can be seen that solving the discrete phase shift design problem is at its early stage.  It is still a challenge to derive a low complexity approach while achieving performance close to that of  brute-force search. 
For the conventional optimization method, the greedy algorithm, despite its heuristic nature, might be suitable here as it has a quadratic complexity order by using a linear search at each step. 
Besides, by viewing the desired phase angle as a non-zero element in a sparse vector \cite{J_ly_ad}, sparse signal processing such as Lasso approximation \cite{tibshirani2011regression} and penalty method \cite{kmtsp} can also be applied to handle discrete random variables. On the other hand, although the DQN algorithm of DRL matches the discrete phase problem, it can only provide a feasible solution and has a slow learning rate and unstable learning process. Making DRL more efficient in wireless applications would be an important direction.

\subsection{Handling Mobility of RISs and Users}

For a large-scale data-centric network, since communication service requirements are highly dynamic and imbalanced among users, it is usually inefficient to deploy RISs at fixed locations. 
To improve network coverage and serve remote nodes, RISs can be deployed on autonomous systems such as unmanned aerial vehicles or unmanned ground vehicles for providing flexible channel reconfigurations. 
Furthermore, the locations of users may also dramatically change over time in emerging V2X networks. Due to the passive nature of RISs, they cannot send pilot signals to track the movement of the users, especially when the direct links from the BS to users are blocked~\cite{M_Pan21RIS_review}. With a mobile RIS or users, the system performance not only depends on the RIS's or users' locations but also on the trajectory itself. Consequently, the dimension of design variables is significantly increased.

Mathematically, the time-varying phase shift design of a mobile wireless system can be modeled as a high-dimensional dynamic programming problem, in which Q-learning, temporal difference learning and policy iteration algorithms in approximate dynamic programming could provide effective solutions \cite{powell2007approximate}. On the other hand, since the CSI for unvisited places and future time slots are unknown, the prior distribution of channels has to be predetermined via the geometry-based tracing approach. However, as time evolves, the knowledge about the channel distribution should be updated for a better phase shift design. This can be modeled as a partially observable Markov decision process, where the DRL methods can be used to learn the underlying wireless environment while deciding the moving trajectory on the fly. Hence, the state of DRL includes not only the current CSI but also the action from the previous time step. Furthermore, by exploiting the extra partial information (e.g., previous locations and velocities of users or the RIS), the post-decision state algorithm can be used to find an optimized solution in dynamic environments during the training of the DRL model~\cite{J_Yang21DRL_RIS}.

\subsection{Scalability of AI-based Methods}
In AI-based methods, while generic multi-layer perceptrons (MLPs) and convolutional neural networks (CNNs) have been widely used for wireless resource allocations, there are two well-known technical challenges. Firstly, MLPs and CNNs are more difficult to train in large-scale settings than small-scale counterparts. For example, as demonstrated in the beamforming problem \cite{syf_gc}, although the performance of CNNs is near-optimal when trained and tested under a two-user setting, there exists a 18\% performance gap to the classic algorithm when trained and tested under a 10-user setting. Secondly, MLPs are designed for a
pre-defined problem size with fixed input and output dimensions. In the context of an RIS problem, this means that a well-trained MLP for a particular RIS dimension is not applicable to another setting when the number of reflecting elements differ. 

Recent studies have shown that incorporating permutation equivariance property into the neural network architecture can reduce the parameter space, avoid a large number of unnecessary permuted training samples, and most importantly make the neural network generalizable to different problem scales \cite{jsac_syf, gj_twc, li2021heterogeneous, 9541201}. In particular, graph neural networks (GNNs) \cite{jsac_syf, gj_twc} and attention-based transformers \cite{li2021heterogeneous, 9541201} have been shown to possess the permutation equivariance property and have demonstrated superior performance, scalability, and generalization ability in a few wireless resource allocation problems. For instance, in the beamforming problem, a GNN trained with data generated in a setting of 50 users was shown to achieve near optimal testing performance under a much larger setting of 1000 users \cite{jsac_syf}. This result in fact simultaneously solves the two challenges mentioned above (difficulty of training in large-scale setting and generalizability to different settings). Interestingly, permutation equivalence also exists in RIS phase shift design problems since exchanging the channels of two reflecting elements should result in a corresponding permutation of the optimized phase shift design. Therefore, it is expected that GNNs and attention-based transformers would be effective neural network architectures for the RIS design problems as well.

\section{Conclusion}
This paper has reviewed and compared current optimization methods for solving resource allocation problems associated with RISs. 
It has been noted that most of the available methods are tailored to the continuous phase shift constraints, and AI-based methods are emerging as serious contenders. With the principles and properties of different algorithms explained and illustrated, and future challenges analyzed, it is hoped that this paper will facilitate the suitable choice of algorithms for future research problems involving RISs.

\appendices

\bibliographystyle{IEEEtran}

\bibliography{MISOtransmissionSystem}

\end{document}